\documentclass{jov}
\usepackage{graphicx}
\usepackage{color}
\usepackage{tablefootnote}
\usepackage{float}
\usepackage{tabularx,colortbl}
\usepackage{bigstrut}
\usepackage{enumitem}
\usepackage{epstopdf}
\usepackage{array}

\usepackage{soul}

\usepackage[cmex10]{amsmath}

\usepackage{url}
\begin{document}

\title{Perceptual Quality Prediction on Authentically Distorted Images Using a Bag of Features Approach}
\abstract{Current top-performing blind perceptual image quality prediction models are generally trained on legacy databases of human quality opinion scores on synthetically distorted images. Therefore they learn image features that effectively predict human visual quality judgments of inauthentic, and usually isolated (single) distortions. However, real-world images usually contain complex, composite mixtures of multiple distortions. We study the perceptually relevant natural scene statistics of such authentically distorted images, in different color spaces and transform domains. We propose a bag of \textit{feature-maps} approach which avoids assumptions about the \textit{type of distortion(s)} contained in an image, focusing instead on capturing consistencies, or departures therefrom, of the statistics of real world images. Using a large database of authentically distorted images, human opinions of them, and bags of features computed on them, we train a regressor to conduct image quality prediction. We demonstrate the competence of the features towards improving automatic perceptual quality prediction by testing a learned algorithm using them on a benchmark legacy database as well as on a newly introduced distortion-realistic resource called the LIVE In the Wild Image Quality Challenge Database. We extensively evaluate the perceptual quality prediction model and algorithm and show that it is able to achieve good quality prediction power that is better than other leading models.}

\author{Ghadiyaram}{Deepti}
 {Department of Computer Science}
 {University of Texas at Austin, Austin, TX, USA}
 {https://www.cs.utexas.edu/~deepti/}
 {deepti@cs.utexas.edu}
\author{Bovik}{Alan C.}
 {Department of Electrical and Computer Engineering}
 {University of Texas at Austin, Austin, TX, USA}
 {http://live.ece.utexas.edu/bovik.php}
 {bovik@ece.utexas.edu}


\maketitle

\section{Introduction}
Objective blind or no-reference (NR) image quality assessment (IQA) is a fundamental problem of vision science with significant implications for a wide variety of image engineering applications. The goal of an NR IQA algorithm is the following: given an image (possibly distorted) and no other additional information, automatically and accurately predict its level of visual quality as would be reported by an average human subject. Given the tremendous surge in visual media content crossing the Internet and the ubiquitous availability of portable image capture (mobile) devices, an increasingly knowledgeable base of consumer users are demanding better quality images and video acquisition and display services. The desire to be able to control and monitor the quality of images produced has encouraged the rapid development of NR IQA algorithms, which can be used to monitor wired and wireless multimedia services, where reference images are unavailable. They can also be used to improve the perceptual quality of visual signals by employing ``quality-centric'' processing, or to improve the perceptual quality of acquired visual signals by perceptually optimizing the capture process. Such ``quality-aware'' strategies could help deliver the highest possible quality picture content to camera users. 

\subsection{Authentic vs. Inauthentic Distortions} \label{authentic}
Current IQA models have been designed, trained, and evaluated on benchmark human opinion databases such as the LIVE Image Quality Database \cite{live-r2}, the TID databases \cite{tid} \cite{tid13}, the CSIQ database \cite{csiq}, and a few other small databases \cite{ivc}. All of these databases have been developed beginning with a small set of high-quality pristine images (29 distinct image contents in \cite{live-r2} and 25 in \cite{tid} \cite{tid13}), which are subsequently distorted. The distortions are introduced in a controlled manner by the database architects and these distortion databases have three key properties. First, the distortion severities / parameter settings are carefully (but artificially) selected, typically for psychometric reasons, such as mandating a wide range of distortions, or dictating an observed degree of perceptual separation between images distorted by the same process. Second, these distortions are introduced by computing them from an idealized distortion model. Third, the pristine images are of very high quality, and are usually distorted by one of several \emph{single} distortions. These databases therefore contain images that have been impaired by one of a few \emph{synthetically} introduced distortion types, at a level of perceptual distortion chosen by image quality scientists. 

Existing legacy image quality databases have played an important role in the advancement of the field of image quality prediction, especially in the design of both distortion-specific and general-purpose full-reference, reduced-reference, and no-reference image quality prediction algorithms. However, the images in these kinds of databases are generally \emph{inauthentically} distorted. Image distortions digitally created and manipulated by a database designer for the purpose of ensuring a statistically significant set of human responses are not the same as the real-world distortions that are introduced during image capture by the many and diverse types of cameras found in the hands of real-world users. We refer to such images as \emph{authentically distorted}. Some important characteristics of real-world, authentically distorted images captured by na{\"i}ve users of consumer camera devices is that the pictures obtained generally cannot be accurately described by a simple generative model, nor as suffering from single, statistically separable distortions. For example, a picture captured using a mobile camera under low-light conditions is likely to be under-exposed, in addition to being afflicted by low-light noise and blur. Subsequent processes of saving and/or transmitting the picture over a wireless channel may introduce further compression and transmission artifacts. Further, the characteristics of the overall distortion ``load'' of an image will depend on the device used for capture and on the camera-handling behavior of the user, which may induce further nonlinear shake and blur distortions. Consumer-grade digital cameras differ widely in their lens configurations, levels of noise sensitivity and acquisition speed, and in post-acquisition in-camera processing. Camera users differ in their shot selection preferences, hand steadiness, and situational awareness. Overall, our understanding of true, authentic image distortions is quite murky. Such complex, unpredictable, and currently unmodeled mixtures of distortions are characteristic of real-world pictures that are authentically distorted. There currently is not any known way to categorize, characterize, or model such complex and uncontrolled distortion mixtures, and it is certainly unreasonable to expect an image quality scientist to be able to excogitate a protocol for creating authentically distorted images in the laboratory, by synthetically combining controlled, programmed distortions into what must ultimately be regarded as authentically distorted images.

There is a way to create databases of authentically distorted images, which is by acquiring images taken by many casual camera users. Normally, inexpert camera users will acquire pictures under highly varied and often suboptimal illuminations conditions, with unsteady hands, and with unpredictable behavior on the part of the photographic subjects. Such real-world, authentically distorted images exhibit a broad spectrum of authentic quality ``types,'' mixtures, and distortion severities, that defy attempts at accurate modeling or precise description. Authentic mixtures of distortions are even more difficult to model when they interact, creating new agglomerated distortions not resembling any of the constituent distortions. A simple example would be a noisy image that is heavily compressed, where the noise presence heavily affects the quantization process at high frequencies, yielding hard-to-describe, visible compressed noise artifacts. Users of mobile cameras will be familiar with this kind of spatially-varying, hard-to-describe distortion amalgamation.

With an overarching goal to design an efficient blind IQA model that operates on images afflicted by real distortions, we created a challenging blind image quality database called the \textbf{LIVE In the Wild Image Quality Challenge Database}. This new database contains images that were captured using a large number of highly diverse individual mobile devices, including tablets and smartphones to acquire typical real scenes in the U.S and Korea. These images are affected by unknown mixtures of generally occurring multiple interacting authentic distortions of diverse severities \cite{crowdsource, deepti-crowdsource}.

A byproduct of the characteristically different natures of authentically distorted images contained in the LIVE Challenge Database is that the statistical assumptions made in the past regarding distorted images do not hold. For example, statistics-based natural scene models, which are highly regular descriptors of natural high-quality images, are commonly modified to account for distortions using ``generalized'' statistical models. We have found that these models lose their power to discriminate high-quality images from distorted images, when the distortions are authentic (Figs. 2 - 4). This is of great consequence for the burgeoning field of blind IQA model development, where the `quality-aware' statistical features used by top-performing no-reference (blind) image quality prediction models such as BRISQUE \cite{brisque}, BLIINDS \cite{bliinds2}, DIIVINE \cite{diivine}, and Tang \emph{et. al} in \cite{lbiq} \cite{tang-cvpr} are highly successful on the legacy IQA databases \cite{live-r2, tid}. However, as we will show, the reliability of these statistical features, and consequently the performances of these blind IQA models suffer when applied to authentically distorted images contained in the new LIVE Challenge Database. We believe that further development of successful no-reference image and video quality models will greatly benefit by the development of authentic distortion databases, making for more meaningful and relevant performance analyses and comparisons of state-of-the-art IQA algorithms, as well as furthering efforts towards improving our understanding of the perception of picture distortions and building better and more robust IQA models.

Here, we aim to produce as rich a set of perceptually relevant quality-aware features as might better enable the accurate prediction of subjective image quality judgments on images afflicted by complex, real-world distortions. Our ultimate goal is to design a more robust and generic quality predictor that performs well not only on the existing legacy IQA databases (such as the LIVE IQA Database \cite{live-r2} and TID 2013 database \cite{tid13}) containing images afflicted only by single, synthetic distortions, but also delivers superior quality prediction performance on real-world images `in the wild,' i.e., as encountered in consumer image capture devices. 

\section{Motivation behind using Natural Scene Statistics}
\textbf{Natural scene statistics:} Current efficient NR IQA algorithms use natural scene statistics (NSS) models \cite{bovik13} to capture the statistical \textit{naturalness}\footnote{Natural images are not necessarily images of natural environments such as trees or skies. Any natural visible-light image that is captured by an optical camera and is not subjected to artificial processing on a computer is regarded here as a natural image including photographs of man-made objects.} (or lack thereof) of images that are not distorted. NSS models rely on the fact that good quality real-world photographic images\footnote{We henceforth refer to such images as `pristine' images.} that have been suitably normalized follow statistical laws. Current NR IQA models measure perturbations of these statistics to predict image distortions. State-of-the-art NSS-based NR IQA models \cite{brisque, bliinds2, diivine, biqi, lbiq, cdiivine} \cite{niqe} exploit these statistical perturbations by first extracting image features in a normalized bandpass space, then learning a kernel function that maps these features to ground truth subjective quality scores. To date, these feature representations have only been tested on images containing synthetically-applied distortions, and may not perform well when applied on real world images afflicted by mixtures of authentic distortions (Table \ref{tbl:svr}).

\begin{figure}[t]
\begin{center}
\includegraphics[width=2in]{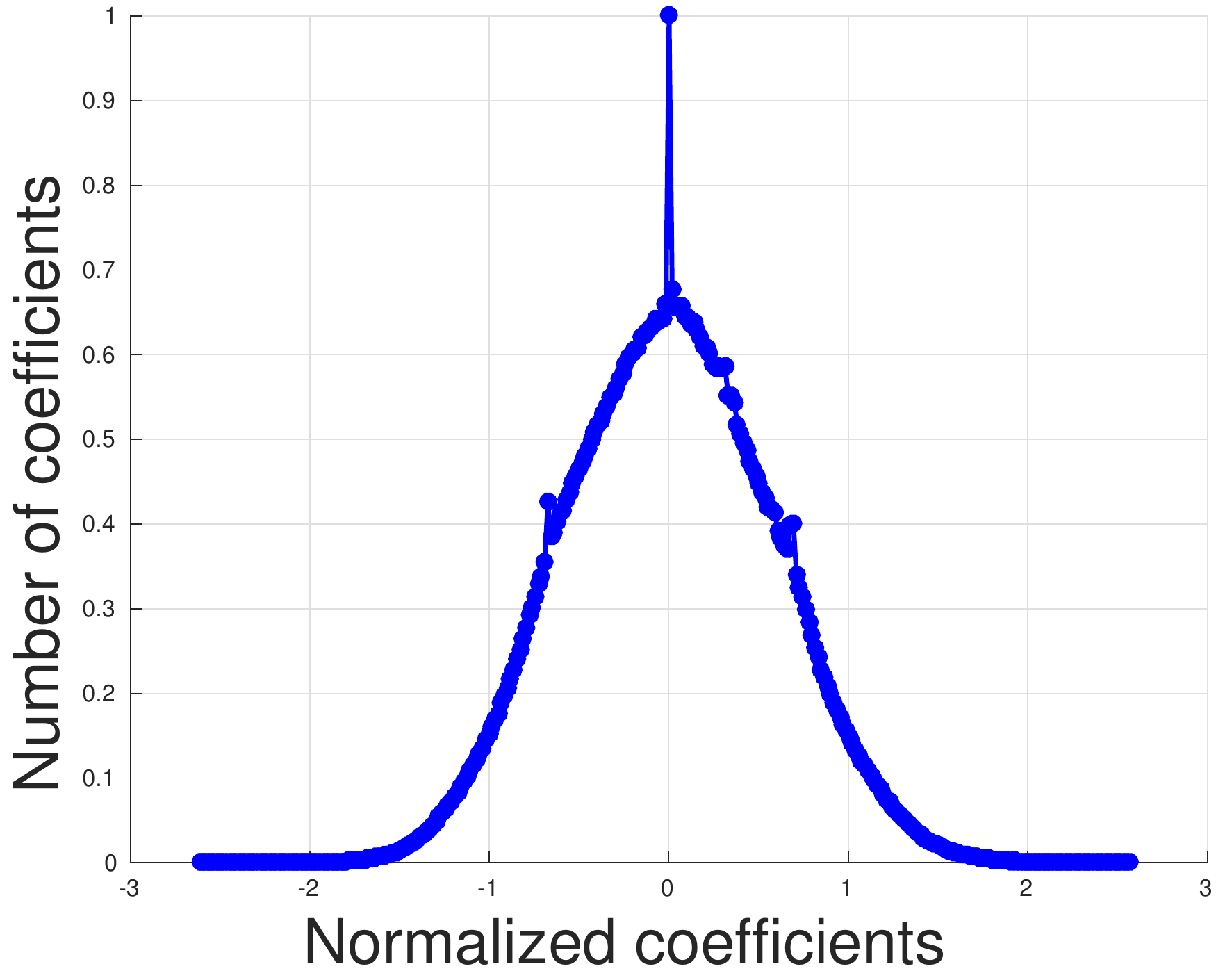}
\caption{Histogram of normalized luminance coefficients of all 29 pristine images contained in the legacy LIVE IQA Database \protect\cite{live-r2}. Notice how irrespective of the wide-variety of image content of the 29 pristine images, their collective normalized coefficients follow a Gaussian distribution (Estimated GGD shape parameter = $2.15$.)}
\label{fig:pristNLCCoeff}
\end{center}
\end{figure} 

\begin{figure}[t]
\begin{center}$
\begin{array}{cccc}
\includegraphics[width=1.5in, height=1.8in]{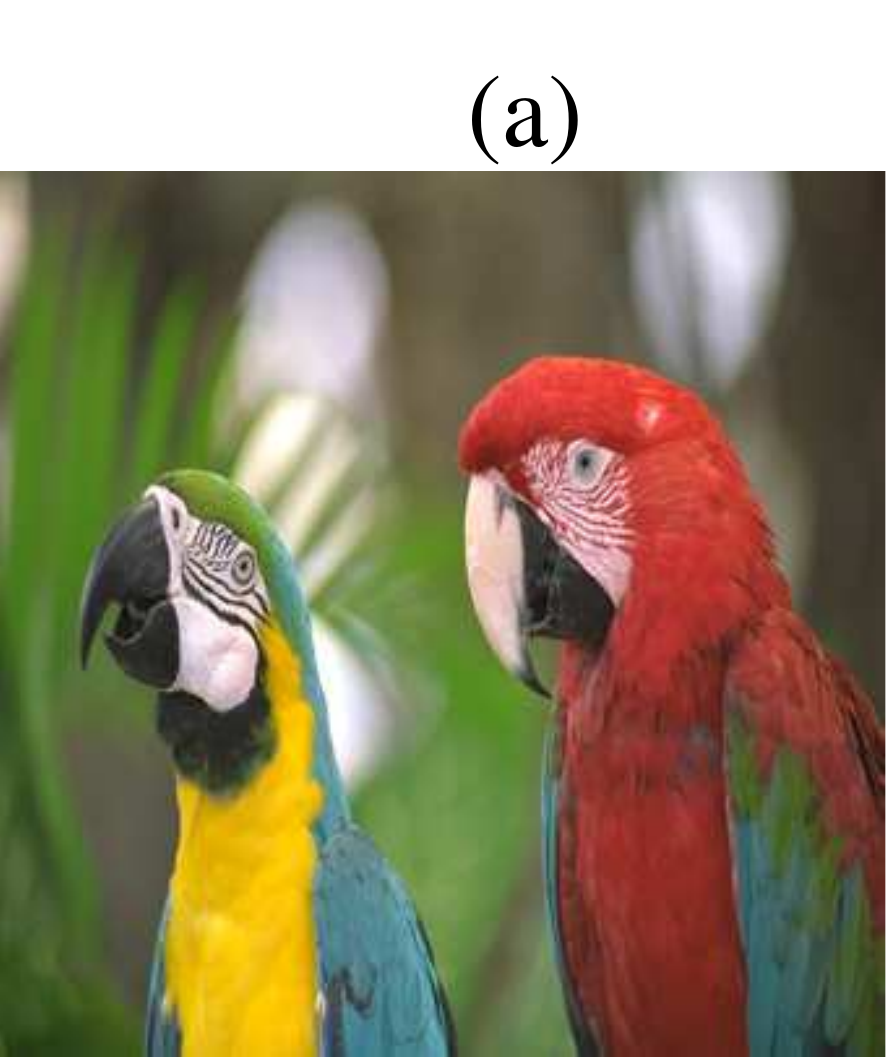} &
\includegraphics[width=1.5in, height=1.8in]{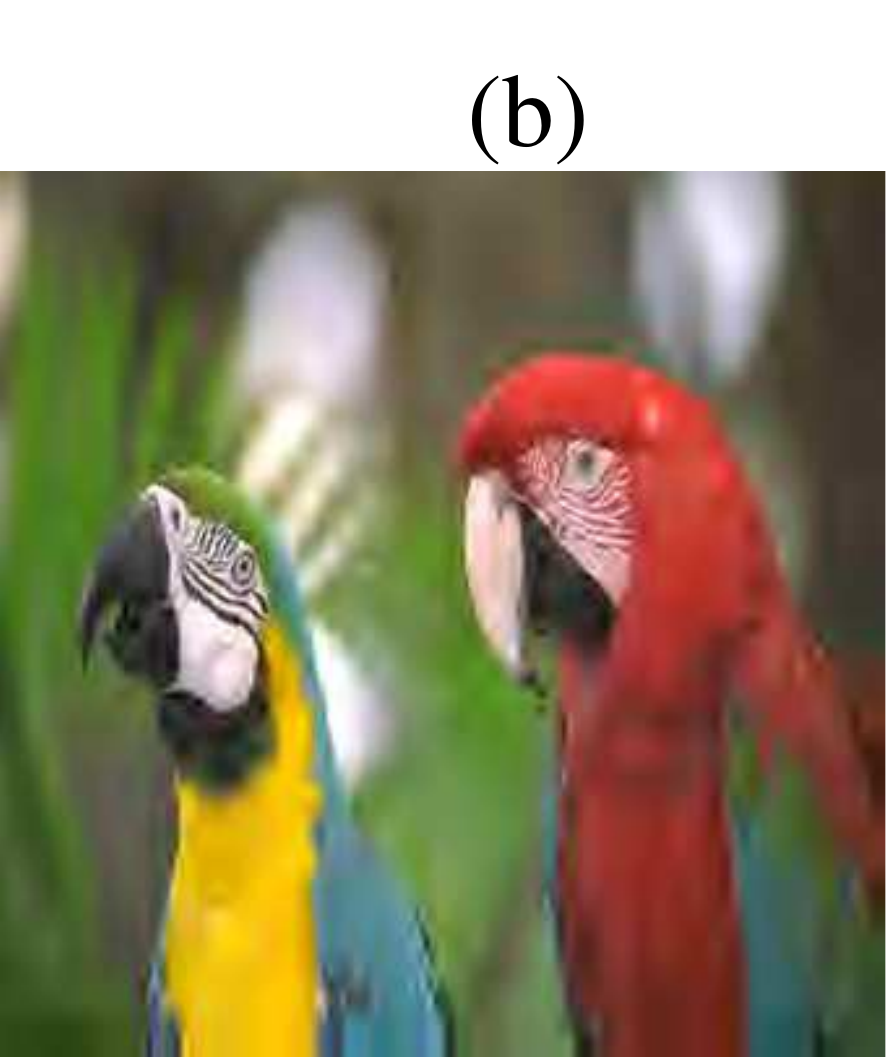} &
\includegraphics[width=1.5in, height=1.8in]{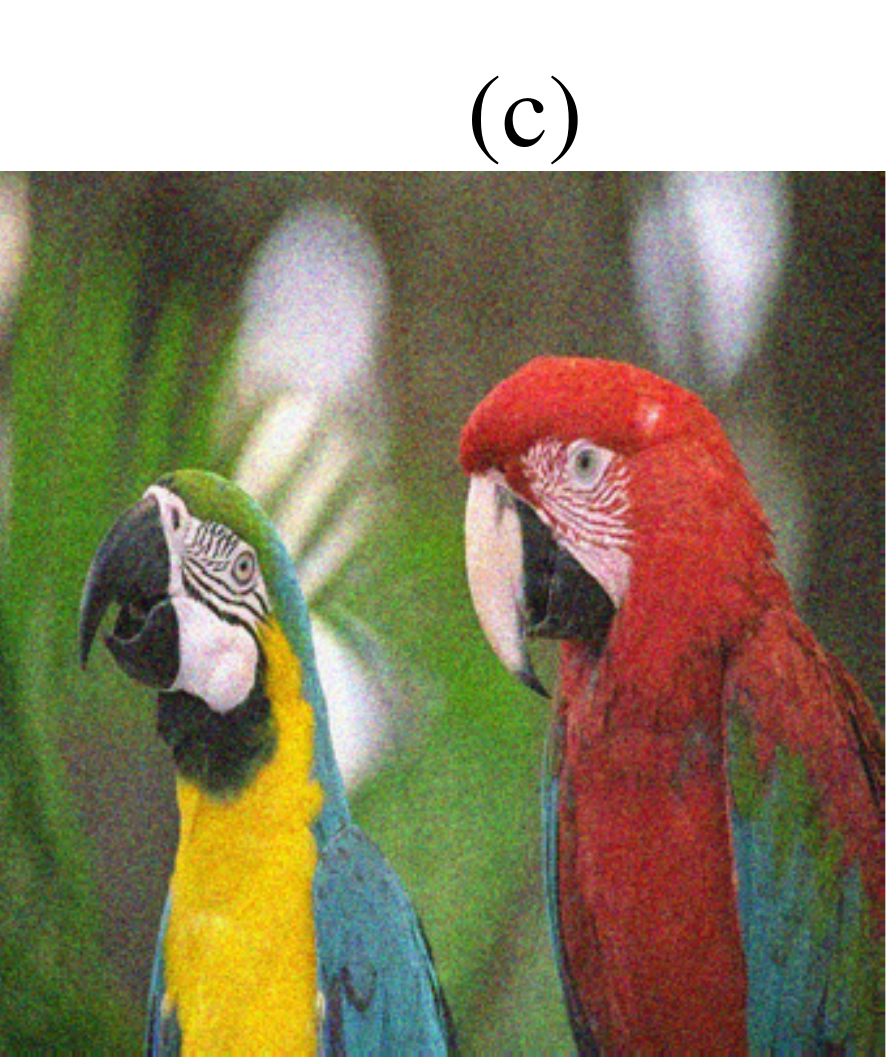} &
\includegraphics[width=1.5in, height=1.8in]{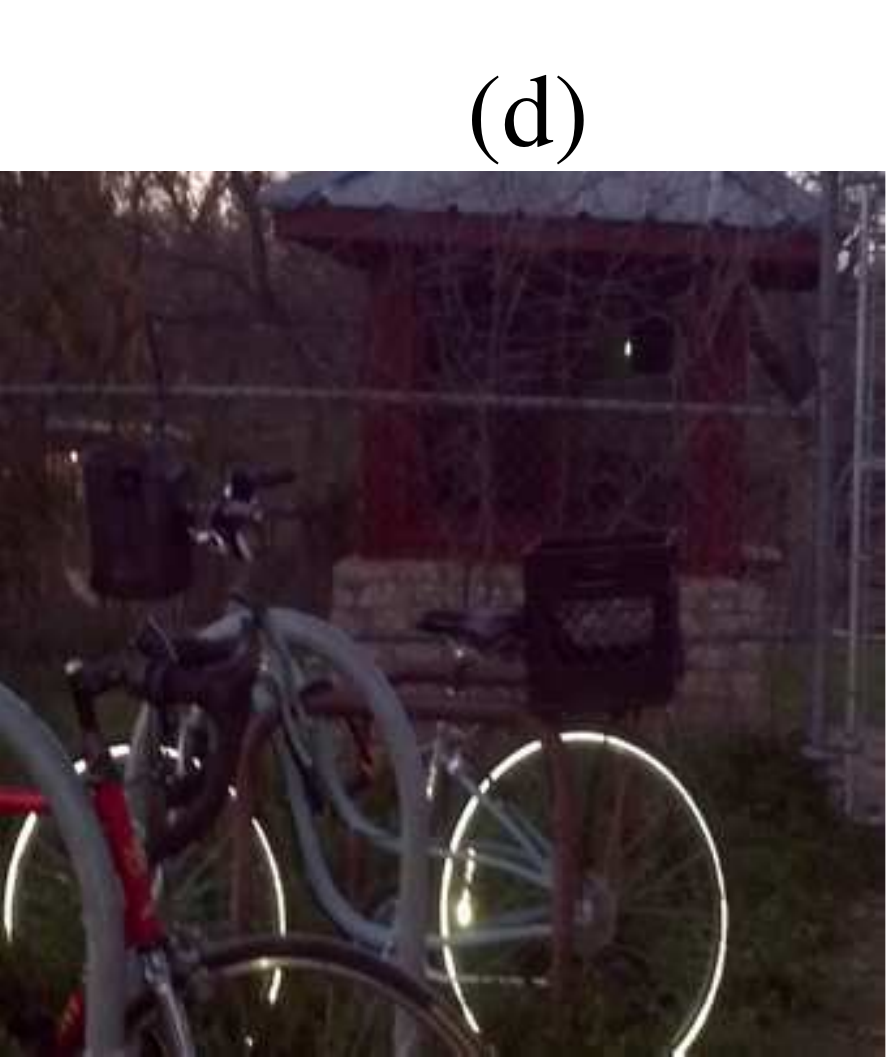}
\end{array}$ 
\caption{{(a) A pristine image from the legacy LIVE Image Quality Database \protect\cite{live-r2} (b) JP2K compression distortion artificially added to (a). (c) White noise added to (a). (d) A blurry image also distorted with low-light noise from the new LIVE In the Wild Image Quality Challenge Database \protect\cite{crowdsource, deepti-crowdsource}.}}
\label{sampleImgs}
\end{center}
\end{figure}

\begin{figure}[t]
\begin{center}
\includegraphics[width=2.9in]{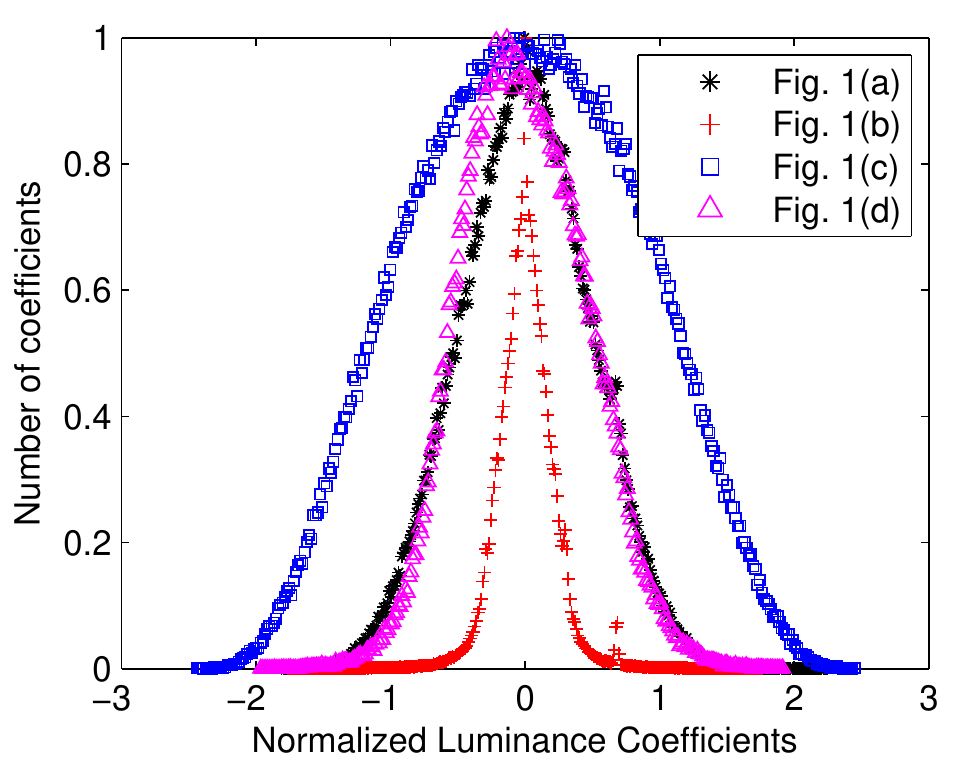}
\caption{{Histogram of normalized luminance coefficients of the images in Figures \ref{sampleImgs}(a) - (d). Notice how each single, unmixed distortion affects the statistics in a characteristic way, but when mixtures of authentic distortions afflict an image, the histogram resembles that of a pristine image. (Best viewed in color). }}
\label{fig:nlcCoeff}
\end{center}
\end{figure} 

\begin{figure}[t]
\begin{center}$
\begin{array}{cc}
\includegraphics[width=3.5in, height=2.25in]{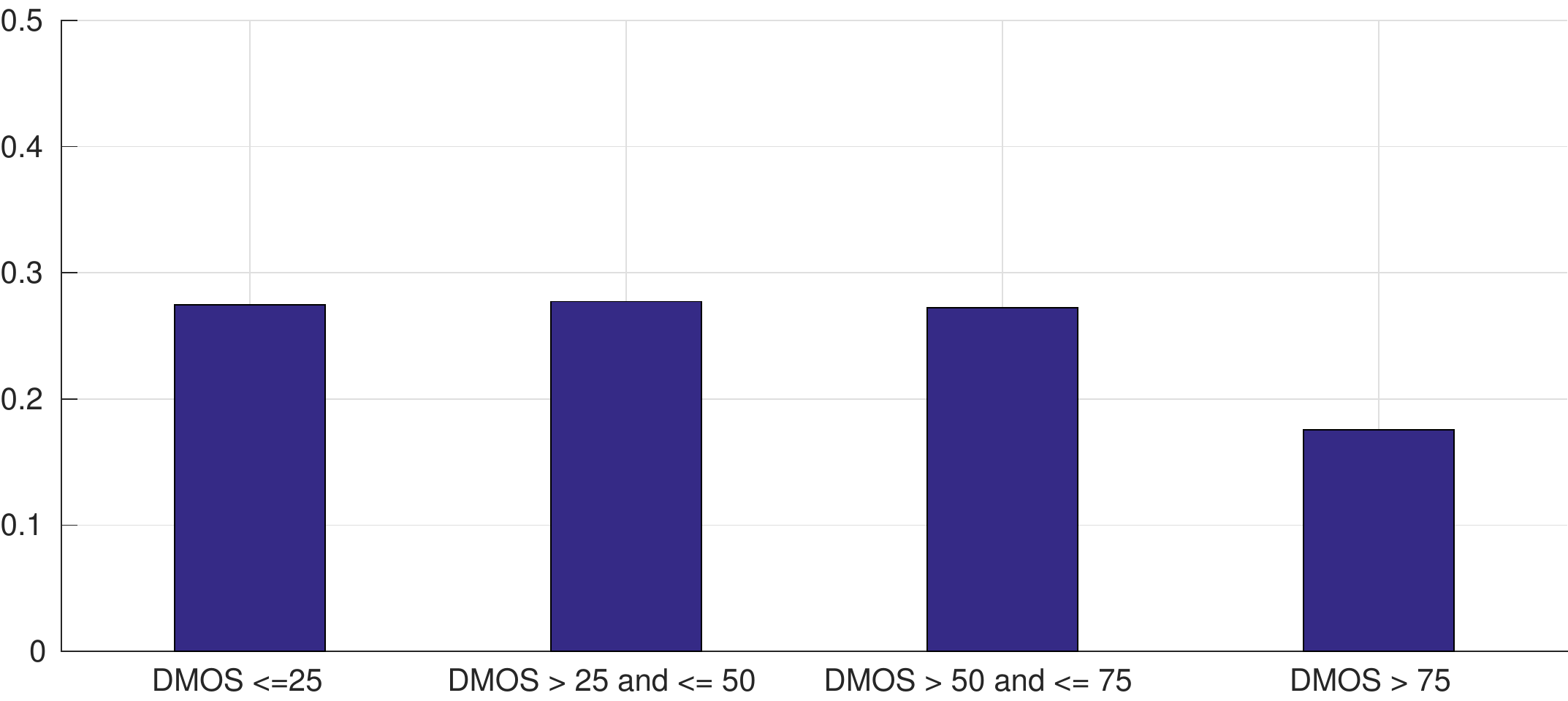} &
\includegraphics[width=3.5in, height=2.25in]{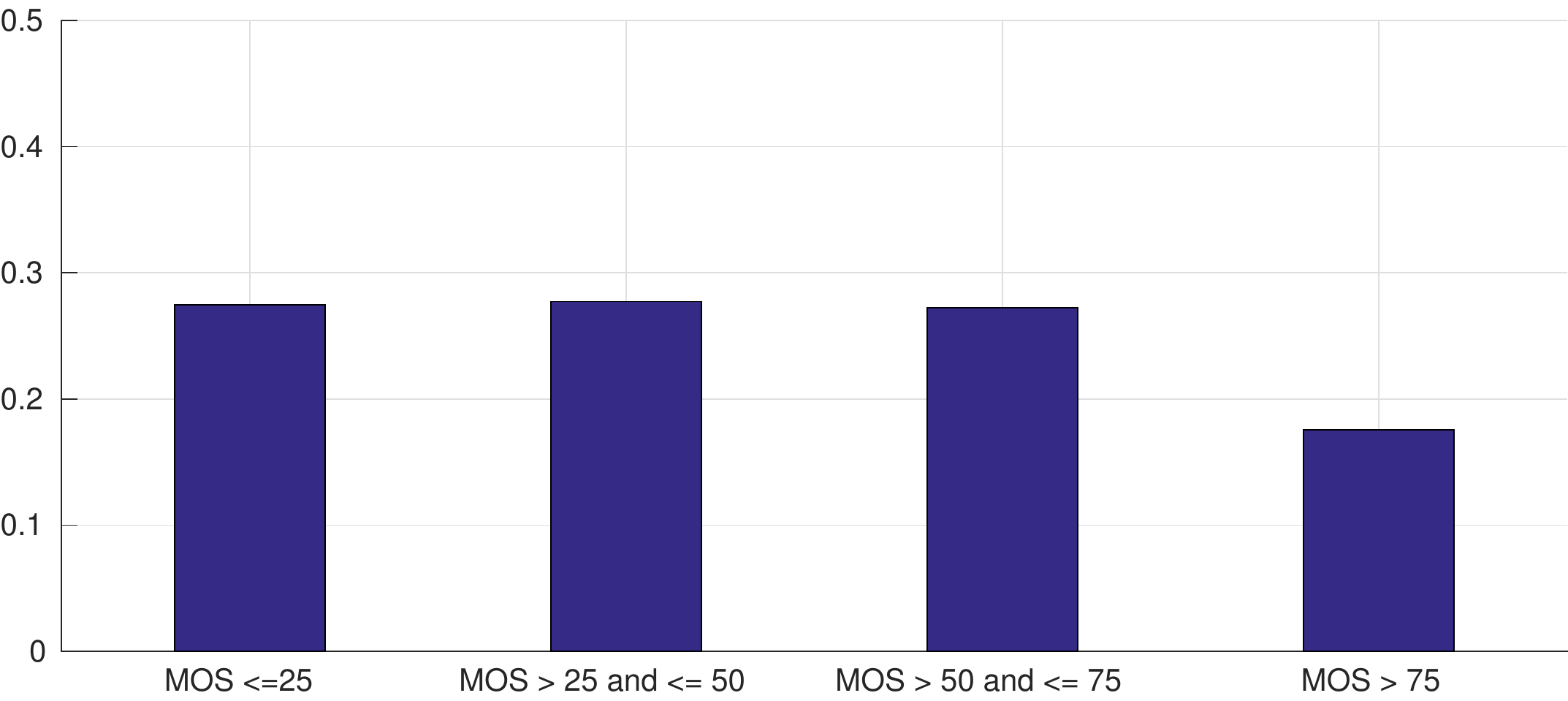}
\end{array}$ 
\caption{{Bar plots illustrating the distribution of the fraction of images from (Left) the legacy LIVE IQA Database and (Right) the LIVE Challenge Database belonging to 4 different DMOS and MOS categories respectively. These histograms demonstrate that the distorted images span the entire quality range in both the databases.}}
\label{fig:MOS_distribution}
\end{center}
\end{figure}

\begin{figure}[t]
\begin{center}$
\begin{array}{cc}
\includegraphics[width=2.25in, height=2.25in]{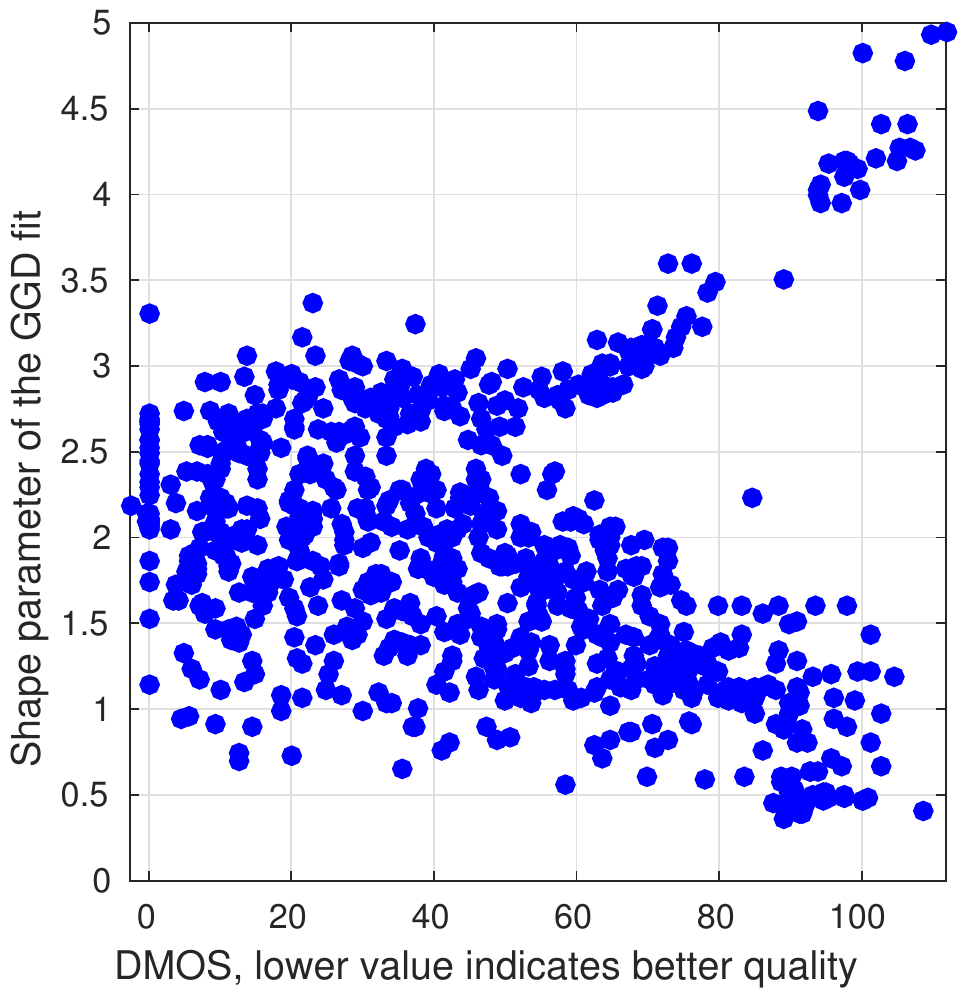} &
\includegraphics[width=2.25in, height=2.25in]{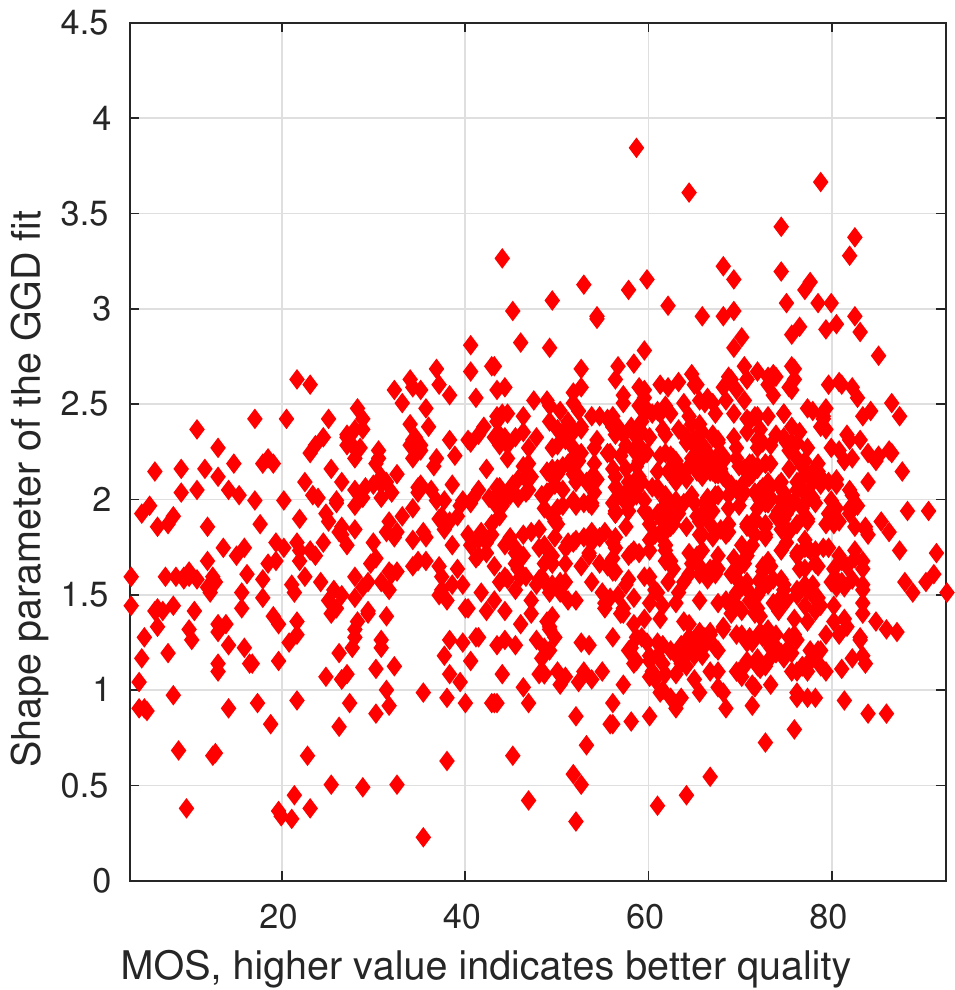}
\end{array}$ 
\caption{{2D scatter plots of subjective quality scores against estimated shape parameters ($\alpha$) obtained by fitting a generalized Gaussian distribution to the histograms of normalized luminance coefficients (NLC) of all the images in (a) the legacy LIVE Database \protect\cite{live-r2} and (b) the LIVE Challenge Database \protect\cite{crowdsource, deepti-crowdsource}.}}
\label{fig:MOS_beta}
\end{center}
\end{figure} 

\begin{figure}[t]
\begin{center}$
\begin{array}{cc}
\includegraphics[width=2.4in, height=2.4in]{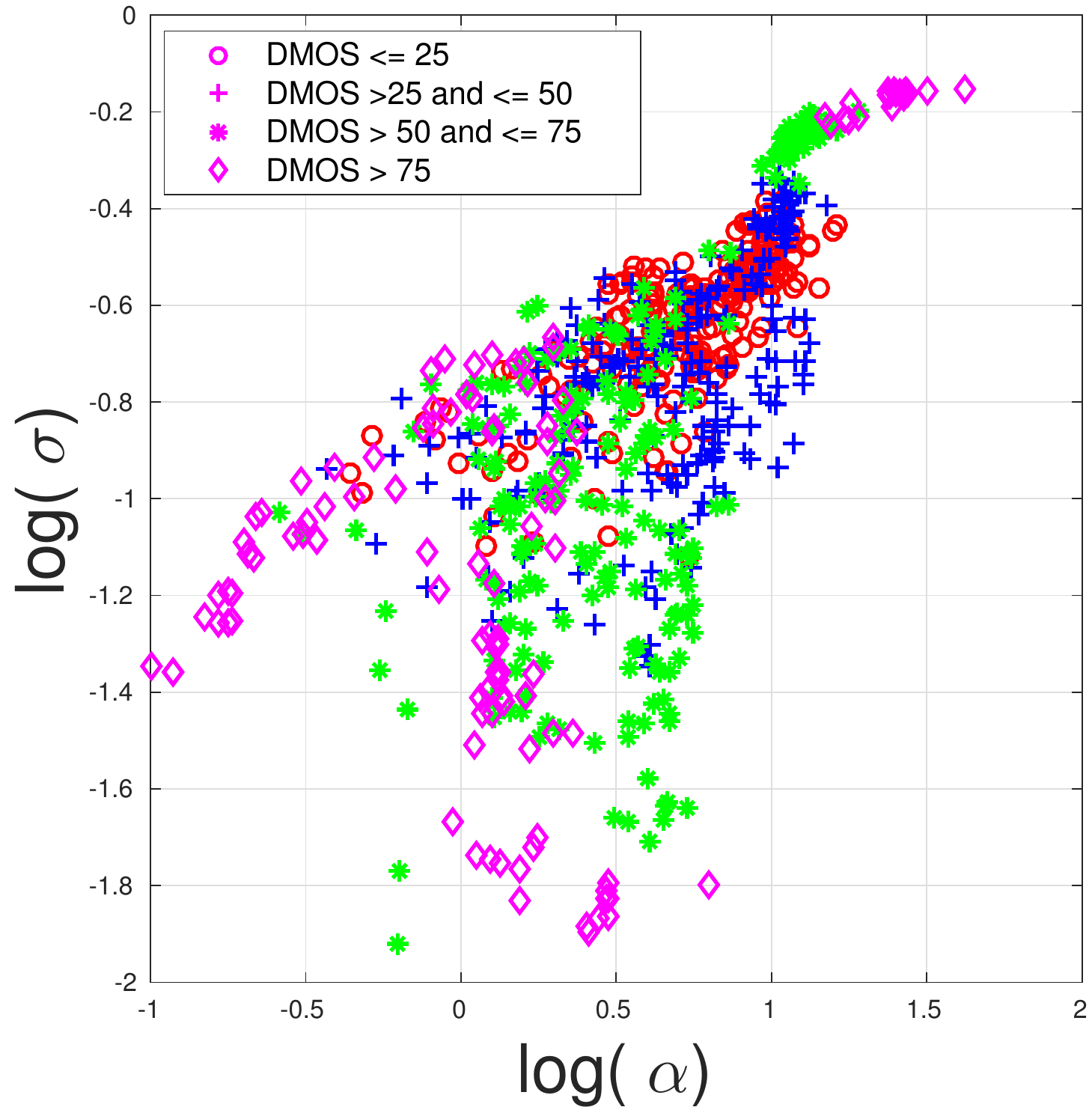} &
\includegraphics[width=2.4in, height=2.4in]{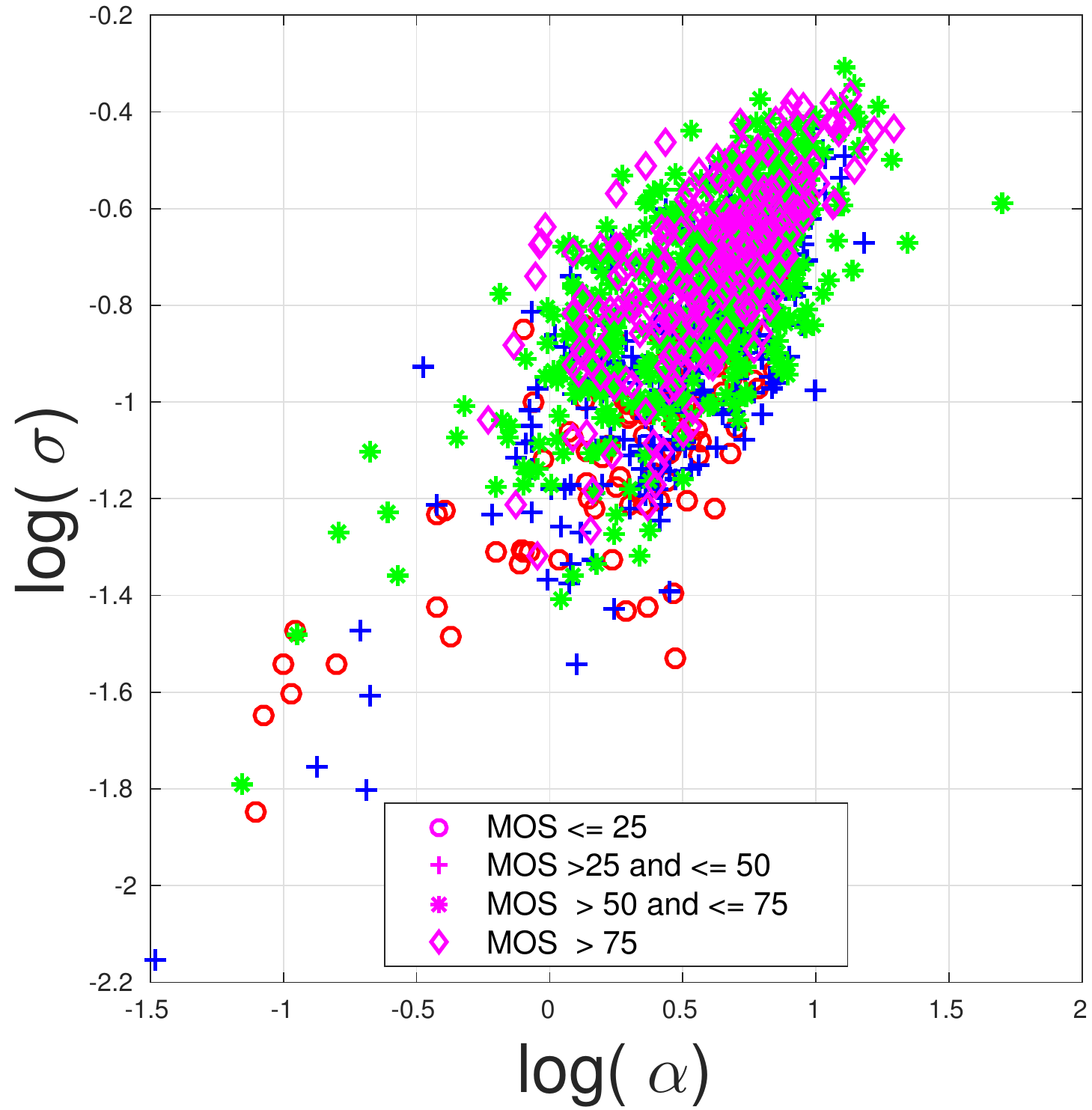}
\end{array}$ 
\caption{{2D scatter plots of the estimated shape and scale parameters obtained by fitting a generalized Gaussian distribution to the histograms of normalized luminance coefficients (NLC) of all the images in (a) the legacy LIVE Database \protect\cite{live-r2} and (b) the LIVE Challenge Database \protect\cite{crowdsource, deepti-crowdsource}. Best viewed in color.}}
\label{fig:beta_sigma}
\end{center}
\end{figure} 

Consider the images in Fig. \ref{sampleImgs}, where images were transformed by a bandpass \emph{debiasing} and \emph{divisive normalization} operation \cite{ruderman}. This normalization process reduces spatial dependencies in natural images. The empirical probability density function (histogram) of the resulting normalized luminance coefficient (NLC) map of the pristine image in Fig. \ref{sampleImgs}(a) is quite Gaussian-like (Fig. \ref{fig:nlcCoeff}). We deployed a generalized Gaussian distribution (GGD) model and estimated its parameters - shape ($\alpha$) and variance ($\sigma^2$) (see below sections for more details). We found that the value of $\alpha$ for Fig. \ref{sampleImgs}(a) is 2.09, in accordance with the Gaussian model of the histogram of its NLC map\footnote{The family of generalized Gaussian distributions include the normal distribution when $\alpha$ = 2 and the Laplacian distribution when $\alpha$ = 1.}. This property is not specific to Fig \ref{sampleImgs}(a), but is generally characteristic of \textit{all} natural images. As first observed in \cite{ruderman} natural, undistorted images of quite general (well-lit) image content captured by any good quality camera may be expected to exhibit this statistical regularity after processing by applying bandpass debiasing and divisive normalization operations. To further illustrate this well-studied phenomenal regularity, we processed 29 pristine images from the legacy LIVE IQA Database \cite{live-r2} which vary greatly in their image content and plotted the collective histogram of the normalized coefficients of all 29 images in Figure \ref{fig:pristNLCCoeff}\footnote{Specifically, we concatenated the normalized coefficients of all the images into a single vector and plotted its histogram.}. The best-fitting GGD model yielded $\alpha = 2.15$, which is again nearly Gaussian. The singular spike at zero almost invariably arises from cloudless sky entirely bereft of objects.

The same property is not held by the distorted images shown in Fig. \ref{sampleImgs}(b) and (c). The estimated shape parameter values computed on those images was $1.12$ and $3.02$ respectively. This deviation from Gaussianity of images containing single distortions has been observed and established in numerous studies on large comprehensive datasets of distorted images, irrespective of the image content. Quantifying these kinds of statistical deviations as learned from databases of annotated distorted images is the underlying principle behind several state-of-the-art objective blind IQA models \cite{brisque, bliinds2, diivine, lbiq, cdiivine, niqe}.

While this sample anecdotal evidence suggests that the statistical deviations of distorted images may be reliably modeled, consider Fig. \ref{sampleImgs}(d), from the new LIVE In the Wild Image Quality Challenge Database\footnote{Throughout this article, we use `LIVE Challenge Database' and `LIVE In the Wild Image Quality Challenge Database' interchangbly.} \cite{crowdsource, deepti-crowdsource}. This image contains an apparent mixture of blur, sensor noise, illumination, and possibly other distortions, all nonlinear and difficult to model. Some distortion arises from compositions of these, which are harder to understand or model. The empirical distribution of its NLC (Fig. \ref{fig:nlcCoeff}) also follows a Gaussian-like distribution and the estimated shape parameter value ($\alpha$) is $2.12$, despite the presence of multiple severe and interacting distortions. As a way of visualizing this problem, we show scatter plots of subjective quality scores against the $\alpha$ values of the best GGD fits to NLC maps of all the images (including the pristine images) in the legacy LIVE database (of synthetically distorted pictures) \cite{live-r2} in Fig. \ref{fig:MOS_beta}(a) and for all the authentically distorted images in the LIVE Challenge database in Fig. \ref{fig:MOS_beta}(b). From Fig. \ref{fig:MOS_beta}(a), it can be seen that most of the images in the LIVE legacy database that have high human subjective quality scores (i.e., low DMOS) associated with them (including the pristine images) have estimated $\alpha$ values close to $2.0$, while pictures having low quality scores (i.e., high DMOS), take different $\alpha$ values, thus are statistically distinguishable from high-quality images. However, Fig. \ref{fig:MOS_beta}(b) shows that authentically distorted images from the new LIVE Challenge database may be associated with $\alpha$ values close to $2.0$, even on heavily distorted pictures (i.e., with low MOS). Fig. \ref{fig:MOS_distribution} plots the distribution of the fraction of all the images in the database that fall into four discrete MOS and DMOS categories\footnote{The legacy LIVE IQA database provides DMOS scores while the LIVE Challenge Database contains MOS scores.}. These histograms show that the distorted images span the entire quality range in both databases and that there is no noticeable skew of distortion severity in either databases that could have affected the results in Figs. \ref{fig:MOS_beta} - \ref{fig:beta_sigma}.

Figure \ref{fig:beta_sigma} also illustrates our observation that authentic and inauthentic distortions affect scene statistics differently. In the case of single inauthentic distortions, it may be observed that pristine and distorted images occupy different regions of this parameter space. For example, images with lower DMOS (higher quality) are more separated from the distorted image collection in this parameter space, making it easier to predict their quality. There is a great degree of overlap in the parameter space among images belonging to the categories `DMOS $<=25$' and `DMOS $>25$ and $<=50$', while heavily distorted pictures belonging to the other two DMOS categories are separated in the parameter space. On the other hand, all the images from the LIVE Challenge database, which contain authentic, often agglomerated distortions overlap to a great extent in this parameter space despite the wide spread of their quality distributions.

Although the above visualizations in Figs. \ref{fig:MOS_beta} and \ref{fig:beta_sigma} were performed in a lower-dimensional space of parameters, it is possible that authentically distorted images could exhibit better separation if modeled in a higher dimensional space of perceptually relevant features. It is clear, however that mixtures of authentic distortions may affect the statistics of images distorted by single, synthetic distortions quite differently. Figures \ref{fig:MOS_beta} and \ref{fig:beta_sigma} also suggest that although the distortion-informative image features used in several state-of-the-art IQA models are highly predictive of the perceived quality of inauthentically distorted images contained in legacy databases \cite{live-r2, tid} (Table \ref{legacy-results}), these features are insufficient to produce accurate predictions of quality on real-world authentically distorted images (Table \ref{tbl:svr}). These observations highlight the need to capture other, more diverse statistical image features towards improving the quality prediction power of blind IQA models on authentically distorted images.

\subsection{Our Contributions and their Relation to Human Vision}
To tackle the difficult problem of quality assessment of images in the wild, we sought to produce a large and comprehensive collection of `quality-sensitive' statistical image features drawn from among the most successful NR IQA models that have been produced to date \cite{cckao}. However, going beyond this, and recognizing that even top-performing algorithms can lose their predictive power on real-world images afflicted by possibly multiple authentic distortions, we also designed a number of statistical features implied by existing models yet heretofore unused, in hopes that they might supply additional discriminative power on authentic image distortion ensembles. Even further, we deployed these models in a variety of color spaces representative of both chromatic image sensing and bandwidth-efficient and perceptually motivated color processing. This large collection of features defined in various complementary perceptually relevant color and transform-domain spaces drives our `feature maps' based approach\footnote{A preliminary version of this work appeared in SPIE \cite{friquee-spie}}. 

Given the availability of a sizeable corpus of authentically distorted images with a very large database of associated human quality judgments, we saw the opportunity to conduct a meaningful, generalized comparative analysis of the quality prediction power of modern `quality-aware' statistical image features defined over diverse transformed domains and color spaces. We thus conducted a discriminant analysis of an initial set of 564 features designed in different color spaces, which, when used to train a regressor, produced an NR IQA model delivering a high level of quality prediction power. We also conducted extensive experiments to validate the proposed model against other top-performing NR IQA models, using both the standard benchmark dataset \cite{live-r2} as well as the new LIVE In the Wild Image Quality Challenge Database \cite{crowdsource, deepti-crowdsource}. We found that \emph{all} prior state-of-the-art NR IQA algorithms\footnote{We could compare only with those algorithms whose code was publicly available.} perform rather poorly on the LIVE Challenge Database, while our perceptually-motivated feature-driven model yielded good prediction performance. These results underscore the need for more representative `quality-aware' NSS features that are predictive of the perceptual severity of authentic image distortions. 

\textbf{Relation to human vision and perception:} The responses of neurons in area V1 of visual cortex perform scale-space orientation decompositions of visual data leading to energy compaction (decorrelation and sparsification) of the data \cite{field}. The feature maps that define our perceptual picture quality model are broadly designed to mimic the processing steps that occur at different stages along the early visual pipeline. While some of the features used have been previously demonstrated to possess powerful perceptual quality prediction capabilities on older, standard `legacy' picture quality databases, we have shown that they perform less effectively on realistic, complex hybrid distortions. As such, we exploit other perceptually relevant features, including a heretofore unused detail (`sigma') feature drawn from current NSS models as well as chromatic features expressed in various perceptually relevant luminance and opponent color spaces. Overall, our feature maps model luminance and chrominance processing in the retina, V1 simple cells, and V1 complex cells, via both oriented and non-oriented multiscale frequency decomposition and divisive contrast normalization processes operating on opponent color channels. The feature maps are also strongly motivated by recent NSS models \cite{brisque, cdiivine, ruderman, srivastava2003} of natural and distorted pictures that are dual to low-level perceptual processing models.

\textbf{Distinction from other machine learning methods:} Although using NSS models for IQA remains an active research area \cite{brisque, cdiivine, diivine, todd, desique}, our bag of features approach goes significantly beyond through the use of a variety of heretofore unexploited perceptually relevant statistical picture features. This places it in distinction with respect to ad hoc machine-learning driven ‘computer vision’ models not founded on perceptual principles \cite{cbiq, lbiq, conv-umd, doermann1}.
\section{Related Work}
\textbf{Blind IQA models:} The development of blind IQA models has been largely devoted to extracting low-level image descriptors that are independent of image content. There are several models proposed in the past that assume a particular kind of distortion and thus extract distortion-specific features \cite{brisque5, brisque6, brisque7, brisque8, brisque9, brisque10, brisque11, chandler1, karam}.

The development of NR IQA models based on natural scene statistics which do not make \emph{a priori} assumptions on the contained distortion is also experiencing a surge of interest. Tang \emph{et al.} \cite{lbiq} proposed an approach combining NSS features along with a very large number of texture, blur, and noise statistic features. The DIIVINE Index \cite{diivine} deploys summary statistics under an NSS wavelet coefficient model. Another model, BLIINDS-II \cite{bliinds2} extracts a small number of NSS features in the DCT domain. BRISQUE \cite{brisque} trains an SVR on a small set of spatial NSS features. CORNIA \cite{cbiq}, which is not an NSS-based model, builds distortion-specific code words to compute image quality. NIQE \cite{niqe} is an unsupervised NR IQA technique driven by spatial NSS-based features, that requires no exposure to distorted images at all. BIQES \cite{biqes} is another recent training-free blind IQA model that uses a model of error visibility across image scales to conduct image quality prediction.

The authors of \cite{conv-umd} use a convolutional neural network (CNN), divide an input image to be assessed into $32 \times 32$ non-overlapping patches, and assign each patch a quality score equal to its source image's ground truth score during training. The CNN is trained on these locally normalized image patches and the associated quality scores. In the test phase, an average of the predicted quality scores is reported. This data augmentation and quality assignment strategy could be acceptable in their work \cite{conv-umd} since their model is trained and tested on legacy benchmark datasets containing single homogeneous distortions \cite{live-r2,tid}. However, our method designs a quality predictor for nonhomogeneous, authentic distortions containing different types of distortions affecting different parts of images with varied severities. Thus, the CNN model and the quality assignment strategy in the training phase and the predicted score pooling strategy in the test phase, as used in \cite{conv-umd} cannot be directly extended to the images in the LIVE Challenge Database. Similarly, the authors of \cite{tang-cvpr} use a deep belief network (DBN) combined with a Gaussian process regressor to train a model on quality features proposed in their earlier work \cite{lbiq}. 

All of these models (other than NIQE) were trained on synthetic, and usually singly distorted images contained in existing benchmark databases \cite{live-r2,tid}. They are also evaluated on the same data challenging their extensibility on images containing complex mixtures of authentic distortions such as those found in the LIVE Challenge Database \cite{crowdsource, deepti-crowdsource}. Indeed, as we show in our experiments, all of the top-performing models perform poorly on the LIVE Challenge Database.

\begin{figure*}[t] 
\begin{center}
\includegraphics[width=8cm]{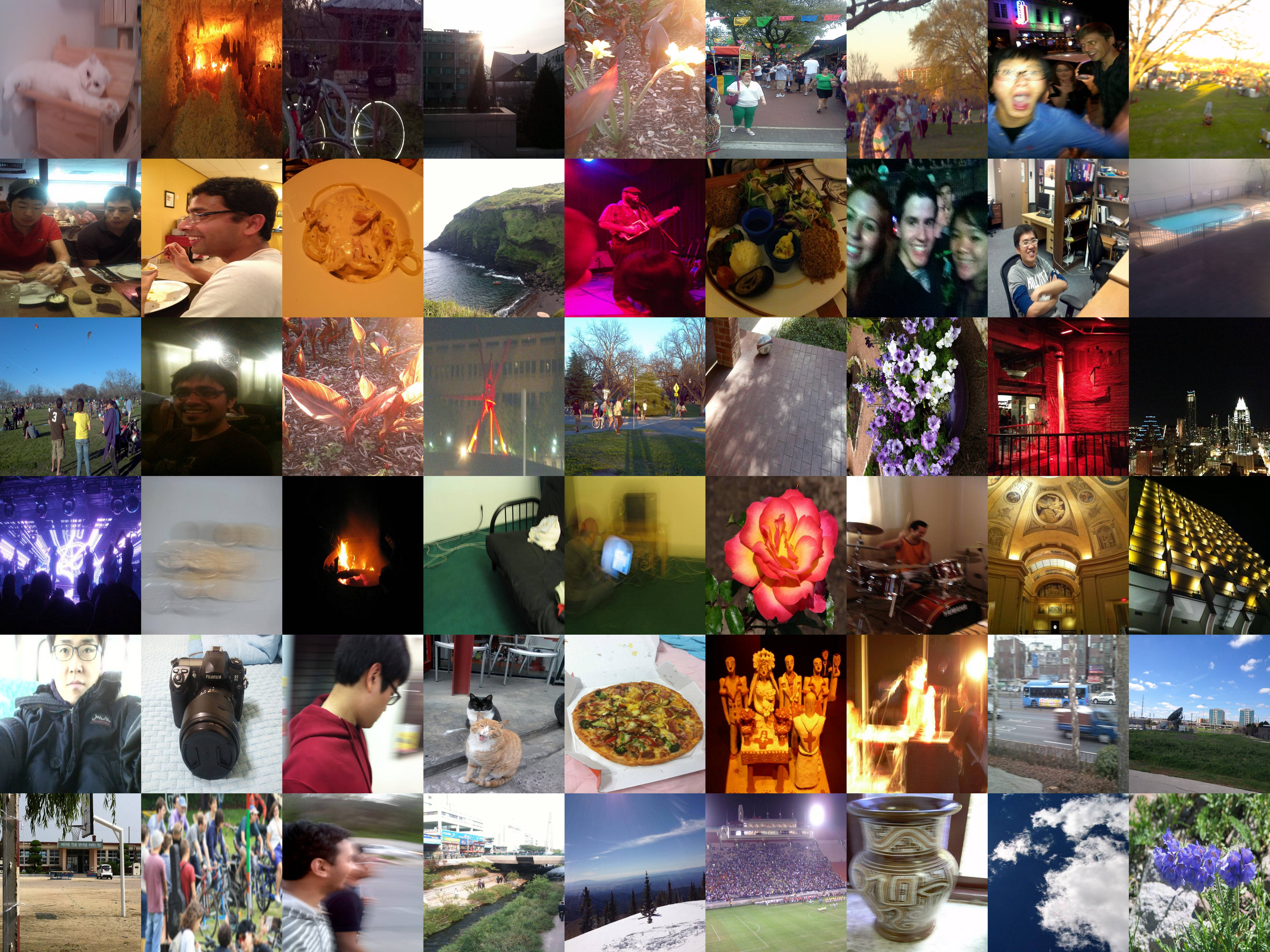}
\caption{{Sample images from the LIVE In the Wild Image Quality Challenge Database \protect\cite{crowdsource, deepti-crowdsource}. These images include pictures of faces, people, animals, close-up shots, wide-angle shots, nature scenes, man-made objects, images with distinct foreground/background configurations, and images without any notable object of interest.}}
\label{fig:challengeImgs}
\end{center}
\end{figure*}

\section{LIVE In the Wild Image Quality Challenge Database}\label{sec:live-challenge}
We briefly describe the salient aspects of the new LIVE Challenge database which helps motivate this work. A much more comprehensive description of this significant effort is given in \cite{crowdsource, deepti-crowdsource}. The new LIVE In the Wild Image Quality Challenge Database \cite{crowdsource} contains $1,163$ images impaired by a wide variety of randomly occurring distortions and genuine capture artifacts that were obtained using a wide-variety of contemporary mobile camera devices including smartphones and tablets. We gathered numerous images taken by many dozens of casual international users, containing diverse distortion types, mixtures, and severities. The images were collected without artificially introducing any distortions beyond those occurring during capture, processing, and storage. 

Figure \ref{fig:challengeImgs} depicts a small representative sample of the images in the LIVE Challenge Database. Since these images are authentically distorted, they usually contain mixtures of multiple impairments that defy categorization into ``distortion types.'' Such images are encountered in the real world and reflect a broad range of difficult to describe (or pigeon-hole) composite image impairments. With a goal to gather a large number of human opinion scores, we designed and implemented an online crowdsourcing system by leveraging Amazon's Mechanical Turk. We used our framework to gather more than 350,000 human ratings of image quality from more than 8,100 unique subjects, which amounts to about 175 ratings on each image in the new LIVE Challenge Database. This study is the world's largest, most comprehensive study of real-world perceptual image quality ever conducted.\\
\textbf{Subject-Consistency Analysis:} Despite the widely diverse study conditions, we observed a very high consistency in users' sensitivity to distortions in images and their ratings. To evaluate subject consistency, we split the ratings obtained on an image into two disjoint equal sets, and computed two MOS values for every image, one from each set. When repeated over 25 random splits, the average linear (Pearson) correlation between the mean opinion scores between the two sets was found to be \textbf{0.9896}. Also, when the MOS values obtained on a fixed set of images (5 gold standard images) via our online test framework were compared with the scores obtained on the same images from a traditional study setup, we achieved a very high correlation of \textbf{0.9851}. Both these experiments highlight the high degree of reliability of the gathered subjective scores and of our test framework.

We refer the reader to \cite{deepti-crowdsource} for details on the content and design of the database, our crowdsourcing framework, and the very large scale subjective study we conducted on image quality. The database is freely available to the public at \url{http://live.ece.utexas.edu/research/ChallengeDB/index.html}

\section{Feature Maps Based Image Quality}\label{sec:friquee}
Faced with the task of creating a model that can accurately predict the perceptual quality of real-world authentically distorted images, an appealing solution would be to train a classifier or a regressor using the distortion-sensitive statistical features that currently drive top-performing blind IQA models. However, as illustrated in Figs. \ref{fig:nlcCoeff} - \ref{fig:beta_sigma}, complex mixtures of authentic image distortions modify the image statistics in ways not easily predicted by these models. They exhibit large, hard to predict statistical variations as compared to synthetically distorted images. Thus, we devised an approach that leverages the idea that different perceptual image representations may distinguish different aspects of the loss of perceived image quality. Specifically, given an image, we first construct several \textit{feature maps} in multiple color spaces and transform domains, then extract individual and collective scene statistics from each of these maps.

Before we describe the types of feature maps that we compute, we first introduce the statistical modeling techniques that we employ to derive and extract features.

\subsection{Statistical Modeling of Normalized Coefficients} \label{modeling}
\textbf{Divisive Normalization:} Wainwright \textit{et al.} \cite{simoncelli}, building on Ruderman's work \cite{ruderman}, empirically determined that bandpass natural images exhibit striking non-linear statistical dependencies. By applying a non-linear \textit{divisive normalization} operation, similar to the non-linear response behavior of certain cortical neurons \cite{heeger-cat}, wherein the rectified linear neuronal responses are divided by a weighted sum of rectified neighboring responses greatly reduces such observed statistical dependencies and tends to guassianize the processed picture data. 

For example, given an image's luminance map $L$ of size $M \times N$, a divisive normalization operation \cite{ruderman} yields a normalized luminance coefficients (NLC) map: \\
\begin{equation} \label{nlc_eqn}
NLC(i,j) = \frac{L(i,j)-\mu(i,j)}{\sigma(i,j) + 1} , 
\end{equation} where 
\begin{equation} \label{mu_eqn}
\mu(i,j) = \sum\limits_{k=-3}^{3} \sum\limits_{l=-3}^{3} w_{k,l}L(i-k,j-l)
\end{equation} and
\begin{equation} \label{sig_eqn}
\sigma(i,j) = \sqrt{\sum\limits_{k=-3}^{3} \sum\limits_{l=-3}^{3} w_{k,l}\left[L(i-k,j-l) - \mu(i-k,j-l)\right]^{2} },
\end{equation}
where $i \in 1,2..M, j \in 1,2..N$ are spatial indices and $w = \{w_{k,l} | k= -3,..., 3, l= -3, ...3\}$ is a 2D circularly-symmetric Gaussian weighting function. 

Divisive normalization by neighboring coefficient energies in a wavelet or other bandpass transform domain similarly reduces of statistical dependencies and gaussianizes the data. Divisive normalization or contrast-gain-control \cite{simoncelli} accounts for specific measured nonlinear interactions between neighboring neurons. It models the response of a neuron as governed by the responses of a pool of neurons surrounding it. Further, divisive normalization models account for the contrast masking phenomena \cite{perception}, and hence are important ingredients in models of distorted image perception.

Most of the feature maps we construct as part of extracting the proposed bag of features are processed using divisive normalization. 

\textbf{Generalized Gaussian Distribution:} Our approach builds on the idea exemplified by observations like those depicted in Fig. \ref{fig:nlcCoeff}, viz., that the normalized luminance or bandpass/wavelet coefficients of a given image have characteristic statistical properties that are predictably modified by the presence of distortions. Effectively quantifying these deviations is crucial to be able to make predictions regarding the perceptual quality of images. A basic modeling tool that we use throughout is the generalized Gaussian distribution (GGD), which effectively models a broad spectrum of (singly) distorted image statistics, which are often characterized by changes in the tail behavior of the empirical coefficient distributions \cite{ggd}. A GGD with zero mean is given by:
\begin{equation} \label{ggd-eqn}
f(x;\alpha,\sigma^2) = \frac{\alpha}{2\beta \Gamma(1/\alpha)} exp \left(-\left(\frac{|x|}{\beta}\right)^\alpha \right),
\end{equation} 
where
\begin{equation} \label{beta}
\beta = \sigma \sqrt{\frac{\Gamma(1/\alpha)}{\Gamma(3/\alpha)}}
\end{equation} 
and $\Gamma(.)$ is the gamma function:
\begin{equation} \label{ggd-gamma}
\Gamma(a) = \int_0^\infty t^{a-1}e^{-t} dt	\; \; \; a > 0.
\end{equation} 

A GGD is characterized by two parameters: the parameter $\alpha$ controls the `shape' of the distribution and $\sigma^2$ controls its variance. A zero mean distribution is appropriate for modeling NLC distributions since they are (generally) symmetric. These parameters are commonly estimated using an efficient moment-matching based approach \cite{ggd} \cite{brisque}.

\begin{figure*}[t] 
\centering
\includegraphics[width=12.0cm]{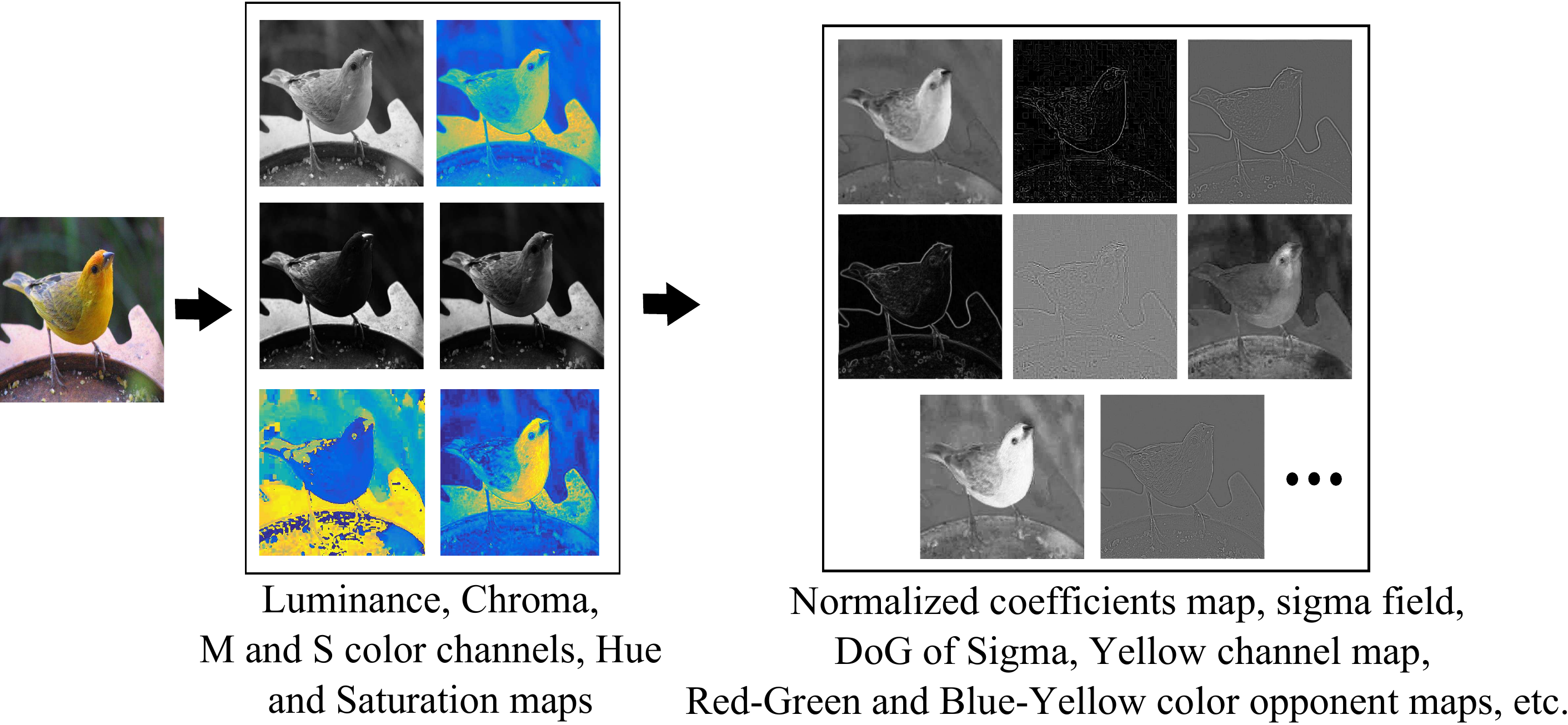}
\caption{{Given any image, our feature maps based model first constructs channel maps in different color spaces and then constructs several feature maps in multiple transform domains on each of these channel maps. Parametric scene statistic features are extracted from the feature maps after performing perceptually significant divisive normalization \protect\cite{ruderman} on them. The design of each feature map is described in detail in later sections.}}
\vspace{0.6cm}
\label{fig:featureMaps}
\end{figure*}

\textbf{Asymmetric Generalized Gaussian Distribution Model:} Additionally, some of the normalized distributions derived from the feature maps are skewed, and are better modeled as following an asymmetric generalized gaussian distribution (AGGD) \cite{aggd}. An AGGD with zero mode is given by: 
\begin{equation} \label{aggd}
f(x;\nu,\sigma_l^2,\sigma_r^2) = \begin{cases}\frac{\nu}{(\beta_l +\beta_r) \Gamma(1/\nu)} exp \left(-\left(\frac{-x}{\beta_l}\right)^\nu \right)  \; \; x<0 \\
\frac{\nu}{(\beta_l +\beta_r) \Gamma(1/\nu)} exp \left(-\left(\frac{x}{\beta_r}\right)^\nu \right)  \; \; x> 0 , \\
\end{cases}
\end{equation}
where
\begin{equation} \label{beta_l}
\beta_l = \sigma_l \sqrt{\frac{\Gamma(1/\alpha)}{\Gamma(3/\alpha)}}
\end{equation} 
\begin{equation} \label{beta_r}
\beta_r = \sigma_r \sqrt{\frac{\Gamma(1/\alpha)}{\Gamma(3/\alpha)}},
\end{equation} 
where $\eta$ is given by:
\begin{equation}\label{eta}
\eta = (\beta_r - \beta_l) \;\frac{\Gamma(2/\nu)}{\Gamma(1/\nu)}.
\end{equation}

An AGGD is characterized by four parameters: the parameter $\nu$ controls the `shape' of the distribution, $\eta$ is the mean of the distribution, and $\sigma_l^2$, $\sigma_r^2$ are scale parameters that control the spread on the left and right sides of the mode, respectively. The AGGD further generalizes the GGD \cite{ggd} and subsumes it by allowing for asymmetry in the distribution. The skew of the distribution is a function of the left and right scale parameters. If $\sigma_l^2 = \sigma_r^2$, then the AGGD reduces to a GGD. All the parameters of the AGGD may be efficiently estimated using the moment-matching-based approach proposed in \cite{aggd}. 

Although pristine images produce normalized coefficients that reliably follow a Gaussian distribution, this behavior is altered by the presence of image distortions. The model parameters, such as the shape and variance of either a GGD or an AGGD fit to the NLC maps of distorted images aptly capture this non-Gaussianity and hence are extensively utilized in our work. Additionally, sample statistics such as kurtosis, skewness, and goodness of the GGD fit, have been empirically observed to also be predictive of perceived image quality and are also considered here. Thus, we deploy either a GGD or an AGGD to fit the empirical NLC distributions computed on different feature maps of each image encountered in \cite{crowdsource, deepti-crowdsource}.

Images are naturally multiscale, and distortions affect image structures across scales. Existing research on quality assessment has demonstrated that incorporating multiscale information when assessing quality produces QA algorithms that perform better in terms of correlation with human perception \cite{bliinds2, ms-ssim}. Hence, we extract these features from many of the feature maps at two scales - the original image scale, and at a reduced resolution (low pass filtered and downsampled by a factor of 2). It is possible that using more scales could be beneficial, but we did not find this to be the case on this large dataset, hence only report scores using two scales. 

\subsection{Feature Maps} \label{sec:featMaps}
Our approach to feature map generation is decidedly a ``Bag of Features'' approach, as is highly popular in the development of a wide variety of computer vision algorithms that accomplish tasks such as object recognition \cite{grauman, bagOfKeyPoints}. However, while our approach uses a large collection of highly heterogeneous features, as mentioned earlier, all of them either have a basis in current models of perceptual processing and/or perceptually relevant models of natural picture statistics, or are defined using perceptually-plausible parametric or sample statistic features computed on the empirical probability distributions (histograms) of simple biologically and/or statistically relevant image features.

\begin{figure}
\begin{center}
\includegraphics[width=11.0cm]{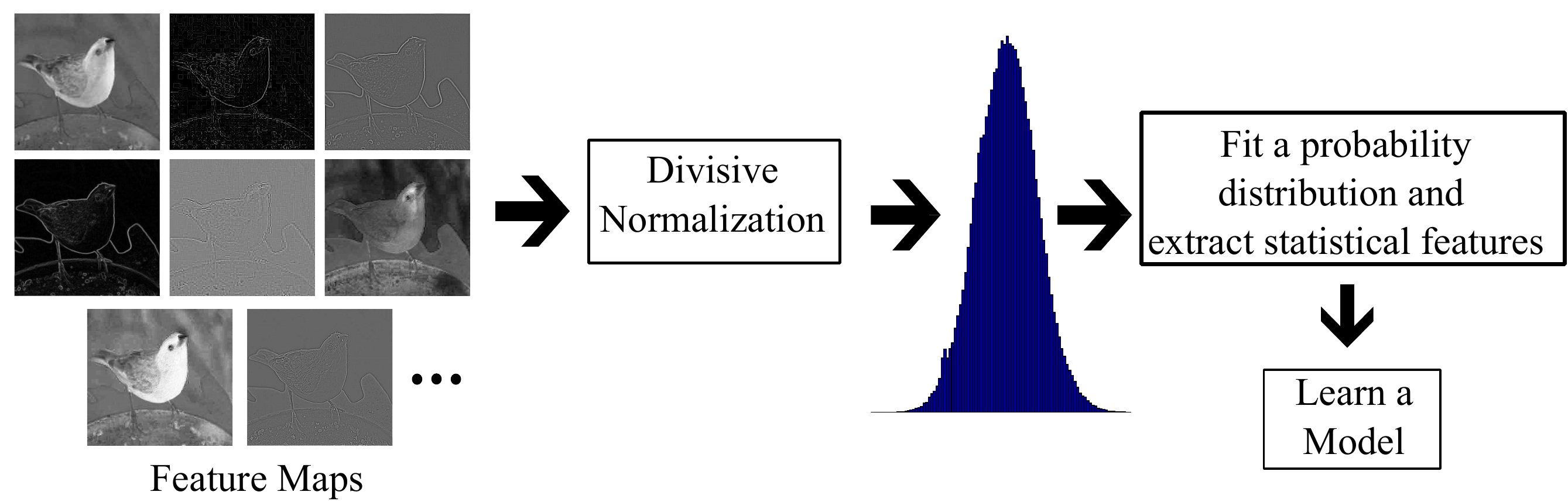}
\caption{{Our proposed model processes a variety of perceptually relevant feature maps by modeling the distribution of their coefficients (divisively normalized in some cases) using either one of GGD (in real or complex domain), AGGD, or wrapped Cauchy distribution, and by extracting perceptually relevant statistical features that are used to train a quality predictor.}}
\label{fig:processFlow}
\end{center}
\end{figure}

We also deploy these kinds of features on a diverse variety of color space representations. Currently, our understanding of color image distortions is quite limited. By using the ``Bag of Features'' approach on a variety of color representations, we aim to capture aspects of distortion perception that are possibly distributed over the different spaces. Figure \ref{fig:featureMaps} schematically describes some of the feature maps that are built into our model, while Fig. \ref{fig:processFlow} shows the flow of statistical feature extraction from these feature maps. 

\subsubsection{Luminance Feature Maps} \label{sec:rgb}
Next we describe the feature maps derived from the luminance component of any image considered.

\textbf{a. Luminance Map:} 
There is considerable evidence that local center-surround excitatory-inhibitory processes occur at several types of retinal neurons \cite{kuffler, bovik13}, thus providing a bandpass response to the visual signal's luminance. It is common to also model the local divisive normalization of these non-oriented bandpass retinal responses, as in \cite{brisque}.

Thus, given an $M \times N \times 3$ image $I$ in RGB color space, its luminance component is first extracted, which we refer to as the $Luma$ map. A normalized luminance coefficient (NLC) map as defined in (\ref{nlc_eqn}) is then computed on it by applying a divisive normalization operation on it \cite{ruderman}. A slight variation from the usual retinal ``contrast signal'' model is the use of divisive normalization by the standard deviation (as defined in (\ref{sig_eqn})) of the local responses rather than by the local mean response. The best-fitting GGD model to the empirical distribution of the NLC map is found \cite{brisque}. Two parameters, ($\alpha, \sigma^2$) are estimated and two sample statistics are computed (kurtosis, skewness) from the empirical distribution over two scales, yielding a total of 8 features. The features may be regarded as essential NSS features related to classical models of retinal processing. 

\begin{figure}[t] 
\begin{center}$
\begin{array}{cc}
\includegraphics[width=1in]{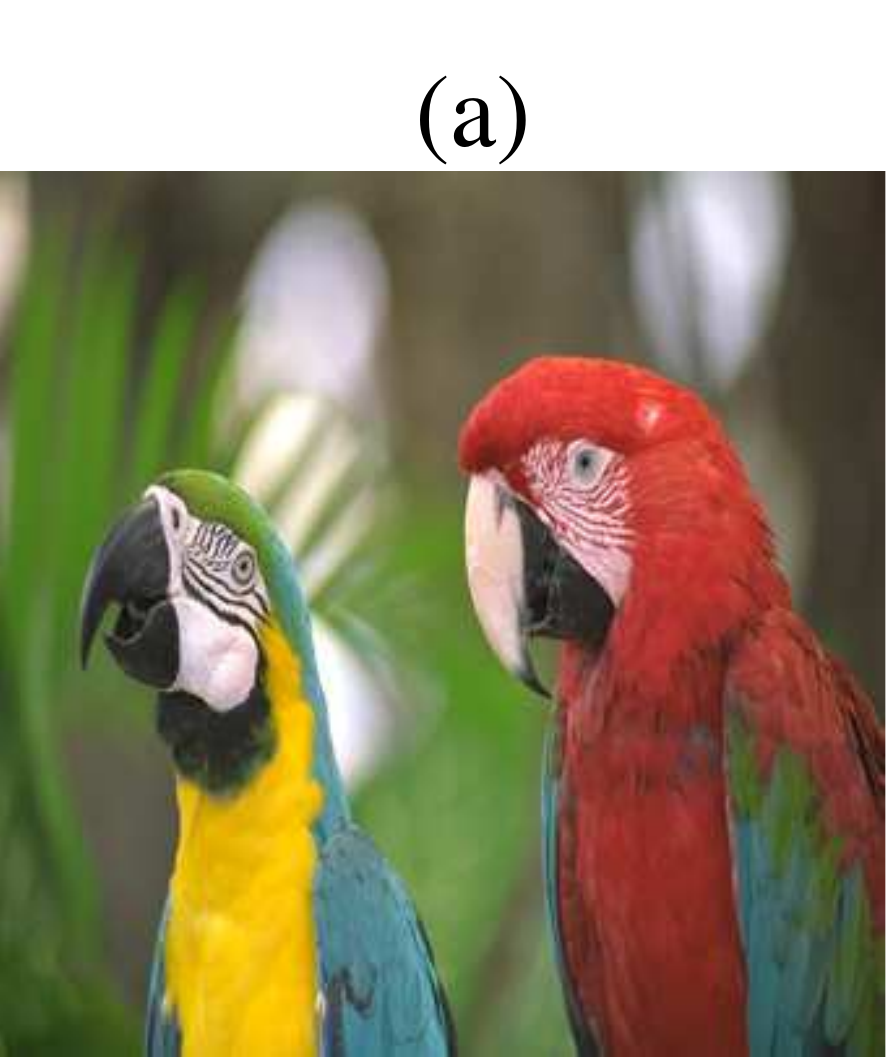} &
\includegraphics[width=1in]{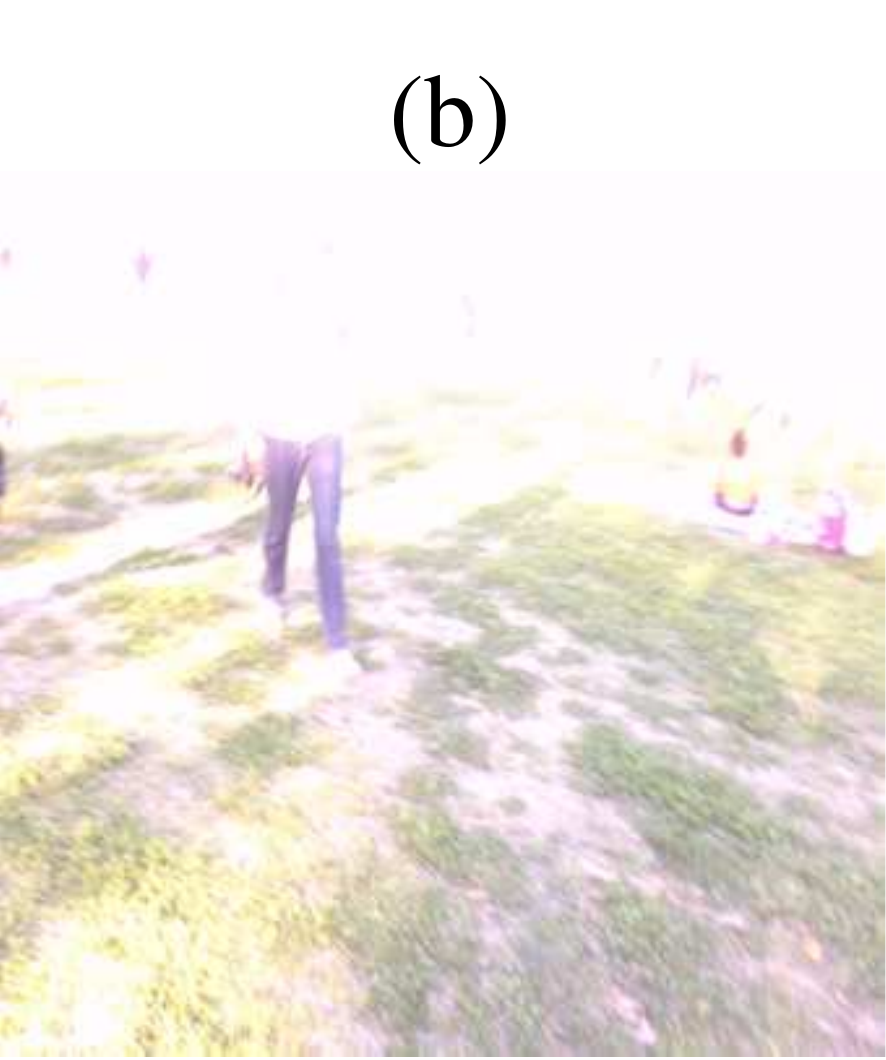} \\
\includegraphics[width=1in]{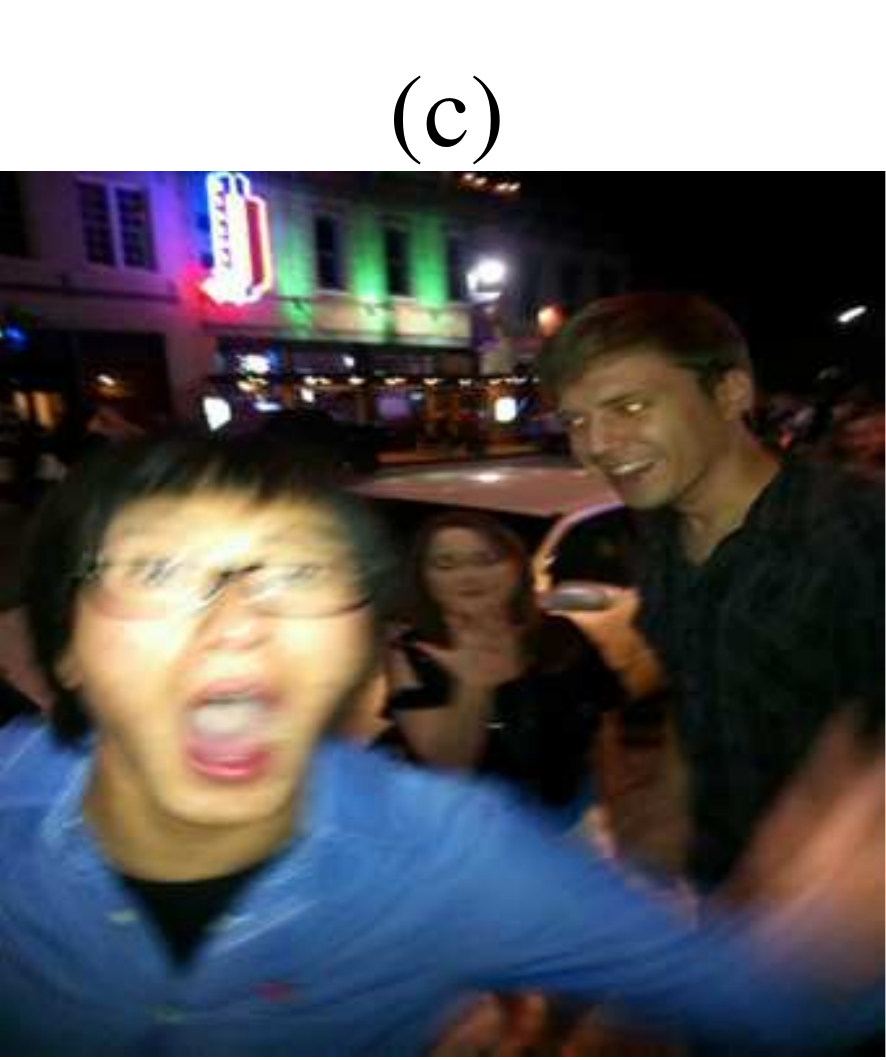} &
\includegraphics[width=1in]{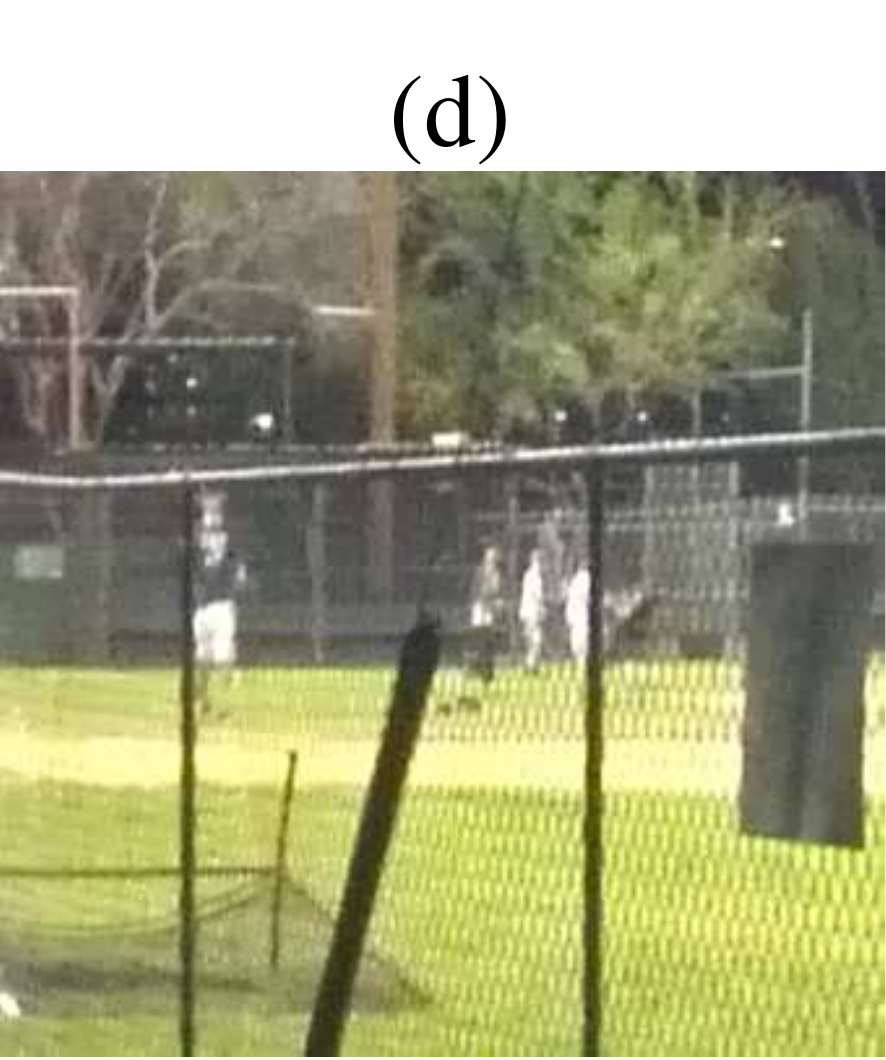}
\end{array}$ 
\caption{{(a) A high-quality image and (b) - (d) a few distorted images from the LIVE Challenge Database \protect\cite{crowdsource, deepti-crowdsource}.} We will use these images in the below sections to illustrate the proposed feature maps and the statistical variations that occur in the presence of distortions.}
\label{fig:imgsForMaps}
\end{center}
\end{figure}

\textbf{b. Neighboring Paired Products:} The statistical relationships between neighborhood pixels of an $NLC$ map are captured by computing four product maps that serve as simple estimates of local correlation. These four maps are defined at each coordinate $(i, j)$ by taking the product of $NLC(i, j)$ with each of its directional neighbors $NLC(i, j+1)$, $NLC(i+1, j)$, $NLC(i+1, j+1)$, and $NLC(i+1, j-1)$. These maps have been shown to reliably obey an AGGD in the absence of distortion \cite{brisque}. A total of 24 parameters (4 AGGD parameters per product map and two sample statistics - kurtosis, skewness) are computed. These features are computed on two scales yielding 48 additional features. These features use the same NSS/retinal model to account for local spatial correlations. 

\textbf{c. Sigma Map:} The designers of existing NSS-based blind IQA models, have largely ignored the predictive power of the sigma field (\ref{sig_eqn}) present in the classic Ruderman model. However, the sigma field of a pristine image also exhibits a regular structure which is disturbed by the presence of distortion. We extract the sample kurtosis, skewness, and the arithmetic mean of the sigma field at $2$ scales to efficiently capture structural anomalies that may arise from distortion. While this feature map has not been used before for visual modeling, it derives from the same NSS/retinal model and is statistically regular. 

\begin{figure}[t] 
\begin{center}$
\begin{array}{cc}
\includegraphics[width=1.9in]{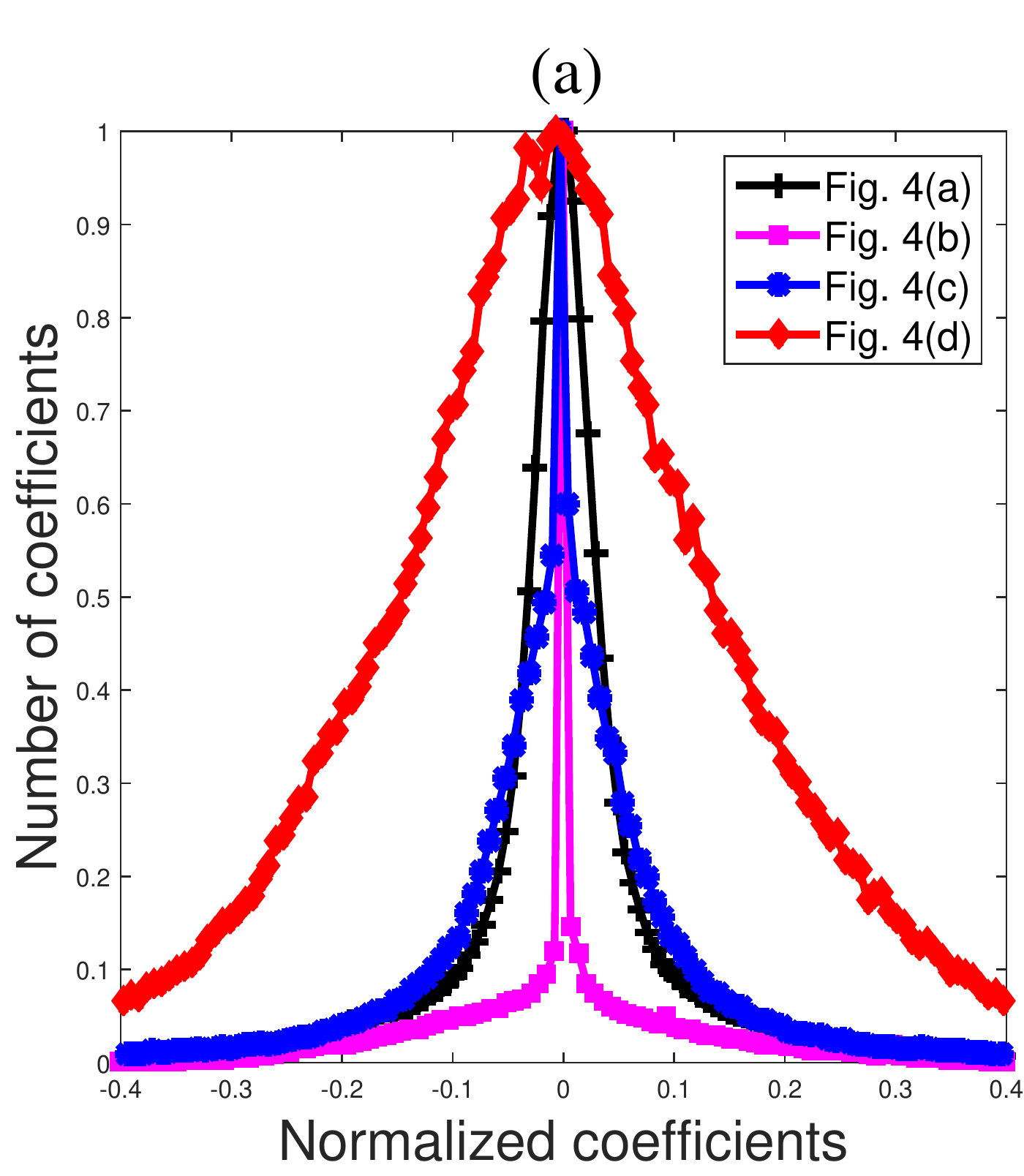} &
\includegraphics[width=1.9in]{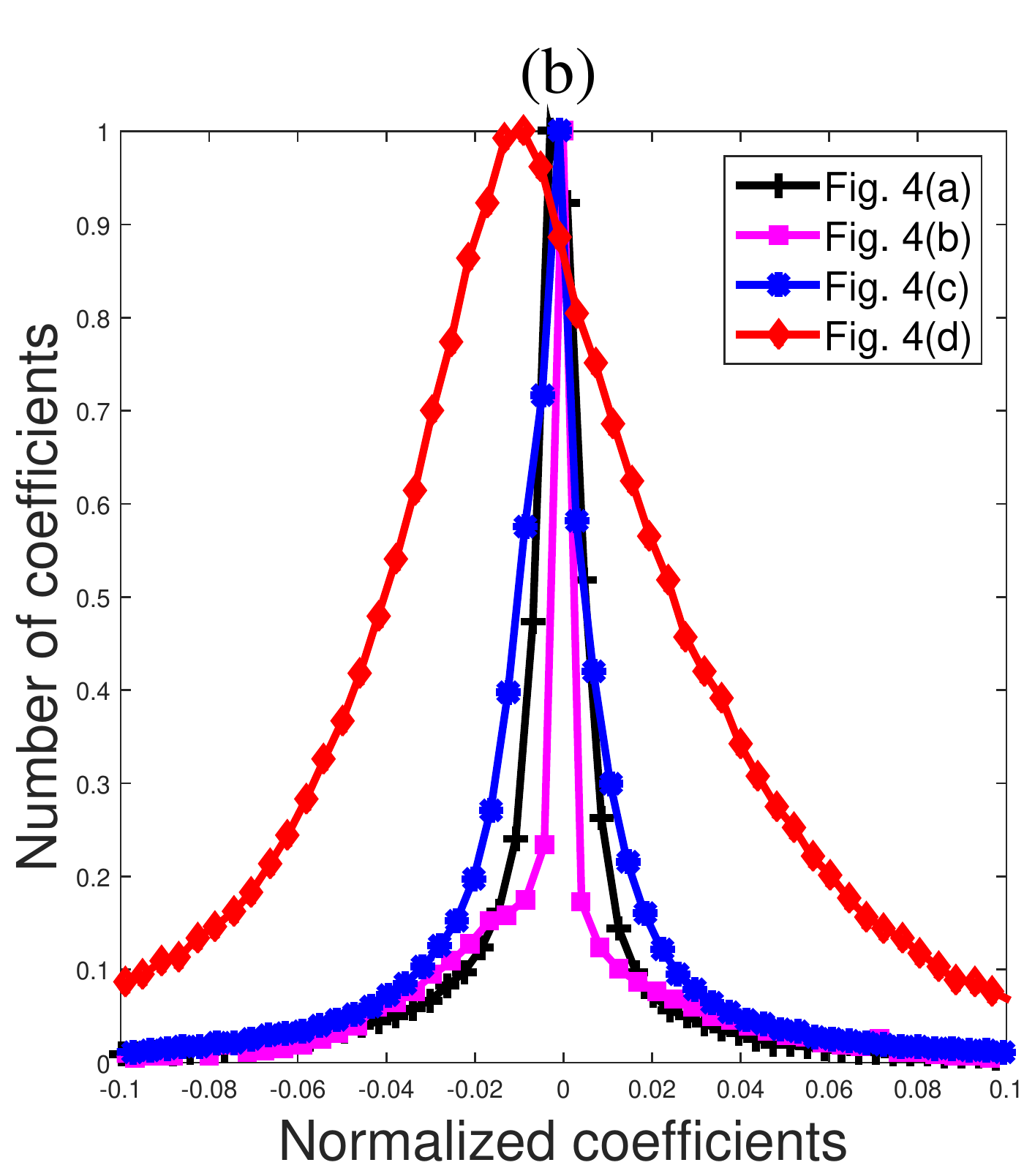} \\
\end{array}$
\caption{{Histogram of normalized coefficients of a) $DoG_{sigma}$ and (b) $DoG^{'}_{sigma}$ of the luminance components of Figures \ref{fig:imgsForMaps} (a) - (d).}}
\label{fig:dogHist}
\end{center}
\end{figure}

\textbf{d. Difference of Gaussian (DoG) of Sigma Map:} Center-surround processes are known to occur at various stages of visual processing, including the multi-scale receptive fields of retinal ganglion cells \cite{cambell}. A good model is the 2D difference of isotropic Gaussian filters \cite{wilson, rodieck}:	
\begin{equation}\label{dog}
DoG  = \frac{1}{\sqrt{2\pi}}\left(\frac{1}{\sigma_1}e^{\frac{-(x^2+y^2)}{2\sigma_{1}^{2}}} - \frac{1}{\sigma_2}e^{\frac{-(x^2+y^2)}{2\sigma_{2}^{2}}}\right) ,
\end{equation}
where $\sigma_2 = 1.5\sigma_1$\footnote{The value of $\sigma_1$ in our implementation was $1.16$}. The mean subtracted and divisively normalized coefficients of the DoG of the sigma field (obtained by applying (\ref{sig_eqn}) on the DoG of the sigma field, denoted henceforth as $\mathbf{DoG_{sigma}}$) of the luminance map of a pristine image exhibits a regular structure that deviates in the presence of some kinds of distortion (Fig. \ref{fig:dogHist}(a)). Features that are useful for capturing a broad spectrum of distortion behavior include the estimated shape, standard deviation, sample skewness and kurtosis. The DoG of the sigma field can highlight conspicuous, `stand-out' statistical features that may particularly affect the visibility of distortions.

We next extract the sigma field of $\mathbf{DoG_{sigma}}$ and denote its mean subtracted and divisively normalized coefficients as $\mathbf{DoG^{'}_{sigma}}$. The sigma field of $\mathbf{DoG_{sigma}}$ is obtained by applying (\ref{sig_eqn}) on $\mathbf{DoG_{sigma}}$. We found that $\mathbf{DoG^{'}_{sigma}}$ also exhibit statistical regularities disrupted by the presence of distortions (Fig. \ref{fig:dogHist}(b)). The sample kurtosis and skewness of these normalized coefficients are part of the list of features that are fed to the regressor.

\textbf{e. Laplacian of the Luminance Map:} A Laplacian image is computed as the downsampled difference between an image and a low-pass filtered version of it. The Laplacian of the luminance map of a pristine image is well-modeled as AGGD, but this property is disrupted by image distortions \cite{zhangZhang}. We therefore compute the Laplacian of each image's luminance map ($Luma$) and model it using an AGGD. This is also a bandpass retinal NSS model, but without normalization. The estimated model parameters $(\nu, \sigma_l^2, \sigma_r^2)$ of this fit are used as features along with this feature map's sample kurtosis and skewness.

\textbf{f. Features extracted in the wavelet domain:}
The next set of feature maps are extracted from a complex steerable pyramid wavelet transform of an image's luminance map. This could also be accomplished using Gabor filters \cite{bovik89} but the steerable pyramid has been deployed quite successfully in the past on NSS-based problems \cite{live-r2, diivine,simoncelli, brisque9}. The features drawn from this decomposition are strongly multiscale and multi-orientation, unlike the other features. C-DIIVINE \cite{cdiivine} is a complex extension of the NSS-based DIIVINE IQA model \cite{diivine} which uses a complex steerable pyramid. Features computed from it enable changes in local magnitude and phase statistics induced by distortions to be effectively captured. One of the underlying parametric probability models used by C-DIIVINE is the wrapped Cauchy distribution. Given an image whose quality needs to be assessed, 82 statistical C-DIIVINE features are extracted on its luminance map using 3 scales and 6 orientations. These features are also used by the learner.

\subsubsection{Chroma Feature Maps}\label{sec:lab}
Feature maps are also defined on the \textit{Chroma} map defined in the perceptually relevant CIELAB color space of one luminance (L*) and two chrominance (a* and b*) components \cite{cielab}. The coordinate L* of the CIELAB space represents color lightness, a* is its position relative to red/magenta and green, and b* is its position relative to yellow and blue. Moreover, the nonlinear relationships between L*, a*, and b* mimic the nonlinear responses of the L, M, and S cone cells in the retina and are designed to uniformly quantify perceptual color differences. \textit{Chroma}, on the other hand, captures the perceived intensity of a specific color, and is defined as follows:
\begin{equation}\label{chroma}
C^{*}_{ab} = \sqrt{{{a^*}^2} + {{b^*}^2}}
\end{equation}
where $a^{*}$ and $b^{*}$ refer to the two chrominance components of any given image in the LAB color space. The chrominance channels contained in the chroma map are entropy-reduced representations similar to the responses of color-differencing retinal ganglion cells.

\begin{figure}[t] 
\begin{center}$
\begin{array}{cc}
\includegraphics[width=1.9in]{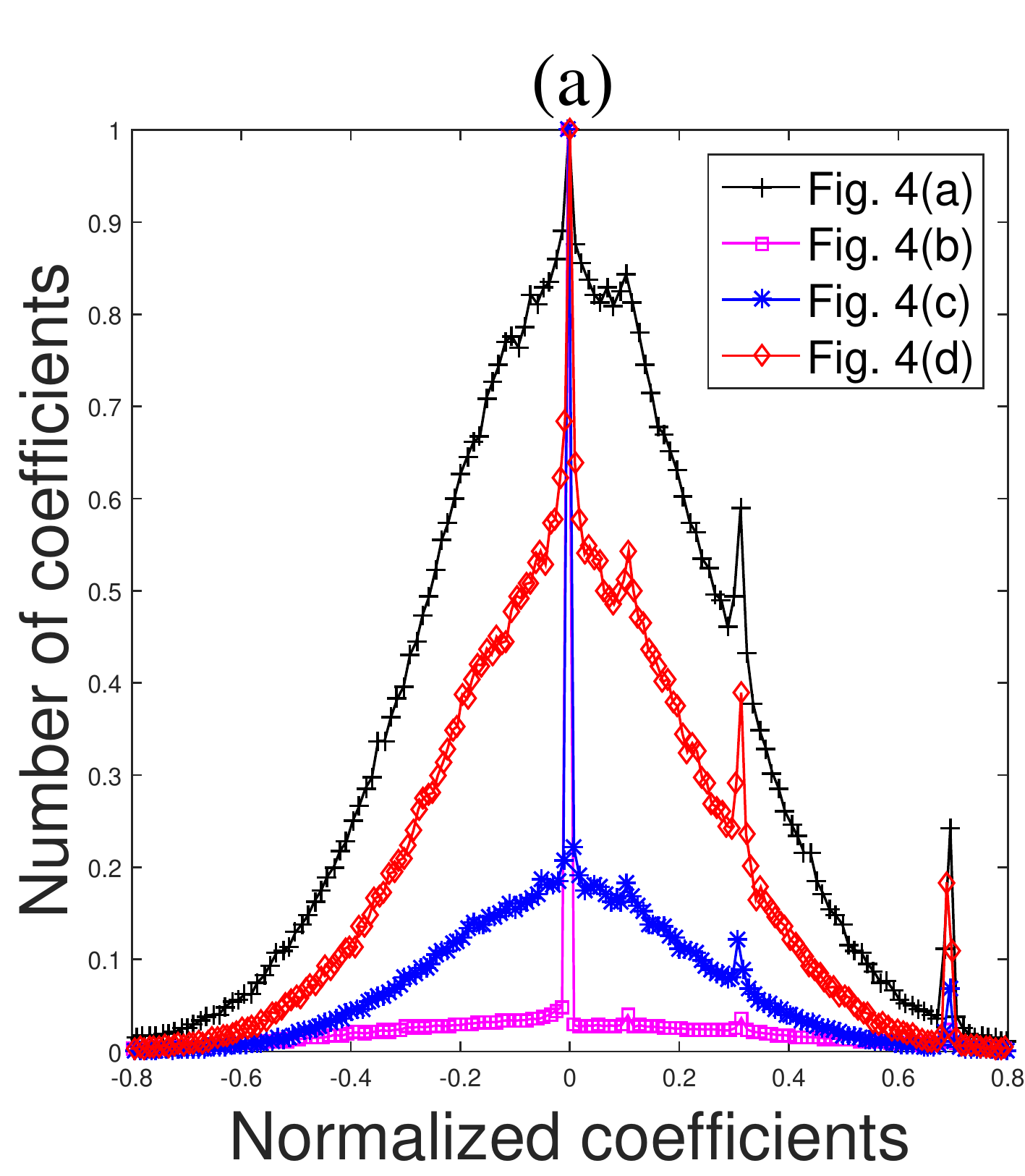} &
\includegraphics[width=1.9in]{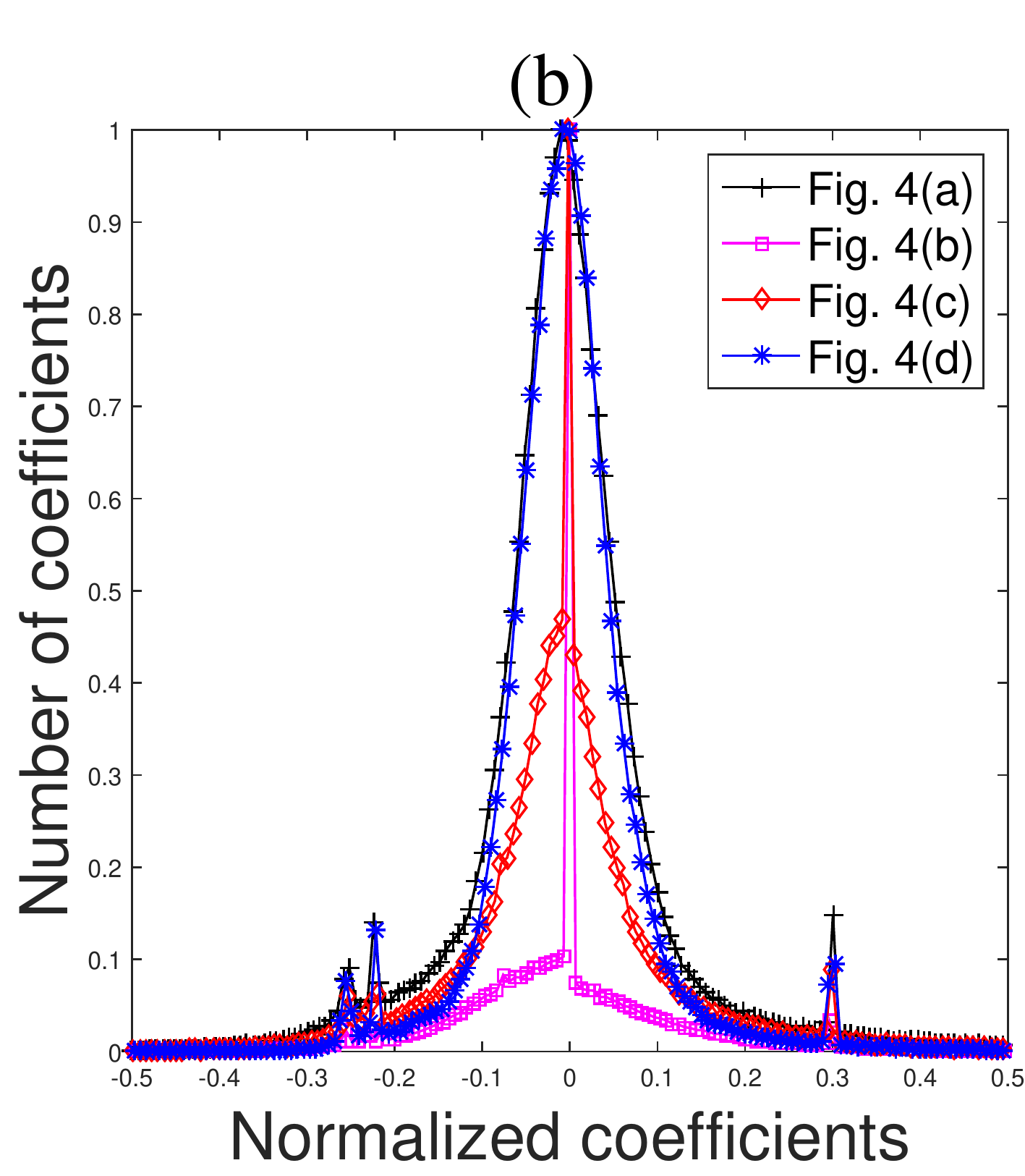} \\
\end{array}$
\caption{{Histogram of normalized coefficients of (a) the $Chroma$ map and (b) $Chroma_{sigma}$ of Fig. \ref{fig:imgsForMaps} (a) - (d).}}
\label{fig:labHist}
\end{center}
\end{figure}

\textbf{g. $\mathbf{Chroma}$ Map:}
The mean subtracted and divisively normalized coefficients of the $Chroma$ map (\ref{chroma}) of a pristine image follow a Gaussian-like distribution, which is perturbed by the presence of distortions (Fig. \ref{fig:labHist} (a)) and thus, a GGD model is apt to capture these statistical deviations. We extract two model parameters -- shape and standard deviation and two sample statistics -- kurtosis and skewness at two scales to serve as image features.

\textbf{h. Sigma field of the $\mathbf{Chroma}$ Map:} We next compute a sigma map (as defined in (\ref{sig_eqn})) of $Chroma$ (henceforth referred to as $Chroma_{sigma}$). The mean subtracted and divisively normalized coefficients of $Chroma_{sigma}$ of pristine images also obey a unit Gaussian-like distribution which is violated in the presence of distortions (Fig. \ref{fig:labHist}(b)). We again use a GGD to model these statistical deviations, estimate the model parameters (shape and standard deviation), and compute the sample kurtosis and skewness at two scales. All of these are used as features deployed by the learner.

Furthermore, as was done on the luminance component's sigma field in the above section, we compute the sample mean, kurtosis, and skewness of $Chroma_{sigma}$. We also process the normalized coefficients of the $Chroma$ map and generate four neighboring pair product maps, the Laplacian, $\mathbf{DOG_{sigma}}$, and $\mathbf{DOG^{'}_{sigma}}$ maps, and extract the model parameters and sample statistics from them. C-DIIVINE features on the $Chroma$ map of each image are also extracted to be used later by the learner.

\subsubsection{LMS Feature Maps}\label{sec:lms}
The LMS color space mimics the responses of the three types of cones in the human retina. Hurvich and Jameson \cite{color-opponency} suggested that the retina contains three types of cone photoreceptors, selectively sensitive to different color mixtures of Long, Medium, and Short wavelengths. They also postulated that each photoreceptor pair has two \emph{physiologically opponent} color members: \emph{red-green, yellow-blue}, in addition to an \emph{achromatic white-black}. Later, Ruderman \emph{et al.} \cite{ruderman-cone} later experimentally gathered cone response statistics and found robust orthogonal decorrelation of the (logarithmic) data along three principal axes, corresponding to one achromatic ($\hat{l}$) and two chromatic-opponent responses ($RG$ and $BY$). 

Denoting $L$, $M$, and $S$ as the three components of LMS color space, the three chromatic-opponent axes are:
\begin{equation}\label{lmsl}
\hat{l} = \frac{1}{\sqrt{3}}\left(\hat{L} + \hat{M} + \hat{S} \right),
\end{equation}
\begin{equation}\label{lmsa}
BY = \frac{1}{\sqrt{6}}\left(\hat{L} + \hat{M} - 2\hat{S} \right),
\end{equation}
\begin{equation}\label{lmsb}
RG = \frac{1}{\sqrt{2}}\left(\hat{L} - \hat{M} \right),
\end{equation}
where $\hat{L}, \hat{M}$, and $\hat{S}$ are the NLCs (\ref{nlc_eqn}) of the logarithmic signals of the L, M, and S components respectively, i.e., 
\begin{equation}\label{logl}
\hat{L}(i,j) = \frac{log L(i,j) - \mu_{L}(i,j)}{\sigma_{L}(i,j) +1} ,
\end{equation}
where $\mu_{L}(i,j)$ is the mean and $\sigma_{L}(i,j) $ is the standard deviation of $log L$, similar to those defined in (\ref{mu_eqn}) and (\ref{sig_eqn}) for $L$. $\hat{M}(i,j)$ and $\hat{S}(i,j)$ are defined in the same manner as (\ref{logl}) from $log M(i,j)$ and $log S(i,j)$ respectively. 

\begin{figure}[t] 
\begin{center}$
\begin{array}{cc}
\includegraphics[width=1.9in]{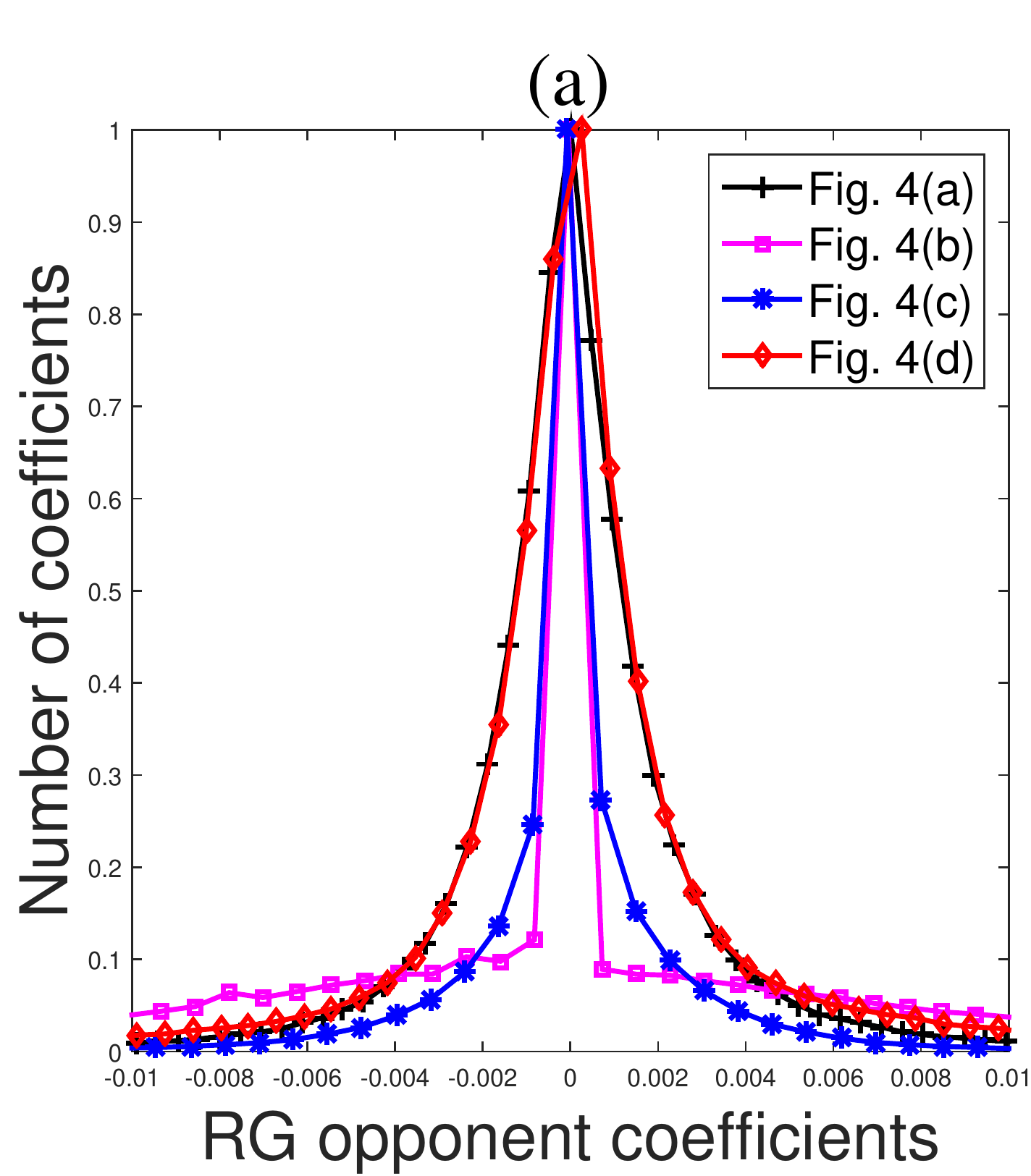} &
\includegraphics[width=1.9in]{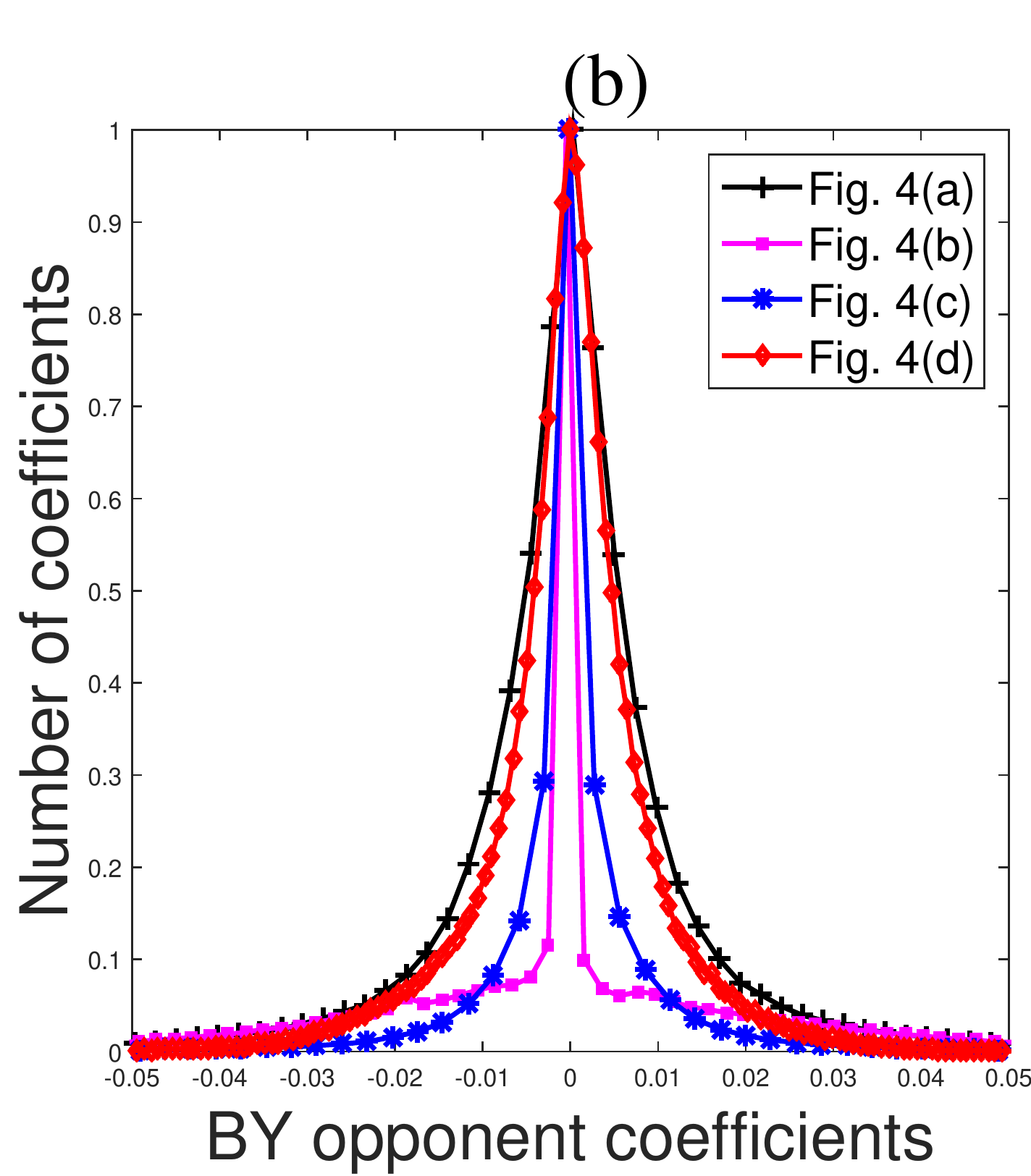} \\
\end{array}$
\caption{{Histogram of color opponent maps of (a) red-green channel ($RG$). (b) blue-yellow channel ($BY$).}}
\label{lms2}
\end{center}
\end{figure}

\textbf{i. Blue-Yellow (BY) and Red-Green (RG) color-opponent maps:}
The marginal distributions of image data projected along each opponent axis follow a Gaussian distribution (Fig. \ref{lms2}). In the presence of distortion, this statistical regularity is perturbed along all three axes (\ref{lmsl}) - (\ref{lmsb}). By projecting each image along the two color opponent axes $RG$ and $BY$, then fitting them with an AGGD model, we are able to capture additional distortion-sensitive features in the form of the model parameters $(\nu, \sigma_l^2, \sigma_r^2)$. We also compute the sample kurtosis and skewness of the color opponent maps $RG$ and $BY$.

\textbf{j. $\mathbf{M}$ and $\mathbf{S}$ Channel Maps:}
After transforming an image into LMS color space, the $M$ and $S$ components are processed as in the previous section and their normalized coefficients are modeled along with their sigma field. We also generate the Laplacian, $\mathbf{DOG_{sigma}}$, and $\mathbf{DOG^{'}_{sigma}}$ feature maps from both $M$ and $S$ channels, and extract model parameters and sample statistics from them. C-DIIVINE features at 3 scales and 6 orientations are also computed on both the channel maps and added to the final list of features.

\begin{table*}[t]
\centering
\scriptsize
\caption{{Summary of different feature maps and the features extracted from them in all three color spaces. The last three columns refer to feature counts from each feature map in each color space and the number in their headings refer to the total number of features in those color spaces.}}
\label{table:feature-all}
\begin{tabular}{|>{\centering\arraybackslash}m{1.7in} | >{\centering\arraybackslash}m{1.5in} | >{\centering\arraybackslash}m{1.1in} | >{\centering\arraybackslash}m{0.9in} | >{\centering\arraybackslash}m{0.3in}|
>{\centering\arraybackslash}m{0.35in}|
>{\centering\arraybackslash}m{0.2in}|}
\hline 
{\textbf{Feature map}} & {\textbf{Color Channels or Spaces}} & {\textbf{Model parameters (derived from GGD (real and complex), AGGD, wrapped Cauchy)}} & {\textbf{Sample statistics}} & {\textbf{Luma} (155)} & {\textbf{Chroma} (163)} & {\textbf{LMS} (240)}\\
\hline
{Yellow color map and its sigma field} & RGB & & {goodness of GGD fit} &  {2} & {0} & {0} \\
\hline
{Red-Green (\ref{lmsb}) and Blue-Yellow (\ref{lmsa}) color opponent maps} & LMS & {shape, left and right standard deviation} & {kurtosis, skewness} & {0} & {0} &{10} \\
\hline
{Neighboring pair product map} & Luminance, Chroma (from LAB space) & {shape, mean, left and right variance} & {kurtosis, skewness} & {48} & {48} & {0}\\
\hline
{Debiased and normalized coefficients} & Luminance, Chroma (from LAB space), M and S (from LMS space)  & {shape, variance} & {kurtosis, skewness} &  {8} & {8} & {16}\\
\hline
{Sigma field} & Luminance, Chroma (from LAB space), M and S (from LMS space) & {shape, variance} & {kurtosis, skewness, mean} &  {6} & {14} & {28} \\
\hline
{$DOG_{sigma}$ and $DOG^{'}_{sigma}$} & Luminance, Chroma (from LAB space), M and S (from LMS space)  & {shape, standard deviation} & {kurtosis, skewness} &  {6} & {6} & {12} \\
\hline
{First Laplacian} & Luminance, Chroma (from LAB space), M and S (from LMS space) & {shape, left and right standard deviations} & {kurtosis, skewness} &  {5} &  {5} &  {10} \\
\hline
{Complex steerable decomposition} & Luminance, Chroma (from LAB space), M and S (from LMS space) & {Model parameters from magnitude and phase coefficients (See \cite{cdiivine})} & {-} &  {82} & {82} & {164}\\
\hline
 \end{tabular}
\vspace{0.3cm}
\end{table*} 

\subsubsection{Statistics from the Hue and Saturation Components} 
We also extract the hue and saturation components of every image in the HSI color space and compute the arithmetic mean and standard deviation of these two components. These four features are also added to the list of features to be considered by the learner. We did examine the bandpass distributions of the HS components, but found that they were redundant with respect to those of other color channels in regards to distortion. Thus, in order to avoid redundancy in our final feature collection, we decided to exclude these from the final feature list.

\subsubsection{Yellow Color Channel Map} Similar to the design of saliency-related color channels in \cite{itti}, we constructed a yellow color channel map of an RGB image $I$, which is defined as follows:
\begin{equation}\label{yellowLedBetter}
Y = \frac{R+G}{2} - \frac{|R-G|}{2} - B,
\end{equation}
where $R$, $G$, and $B$ refer to the red, green, and blue channels respectively. Our motivation for using the yellow channel is simply to provide the learner with direct yellow-light information rather than just B-Y color opponency, which might be relevant to distortion perception, especially on sunlit scenes.  

\begin{figure}[t] 
\begin{center}$
\begin{array}{cc}
\includegraphics[width=1.9in]{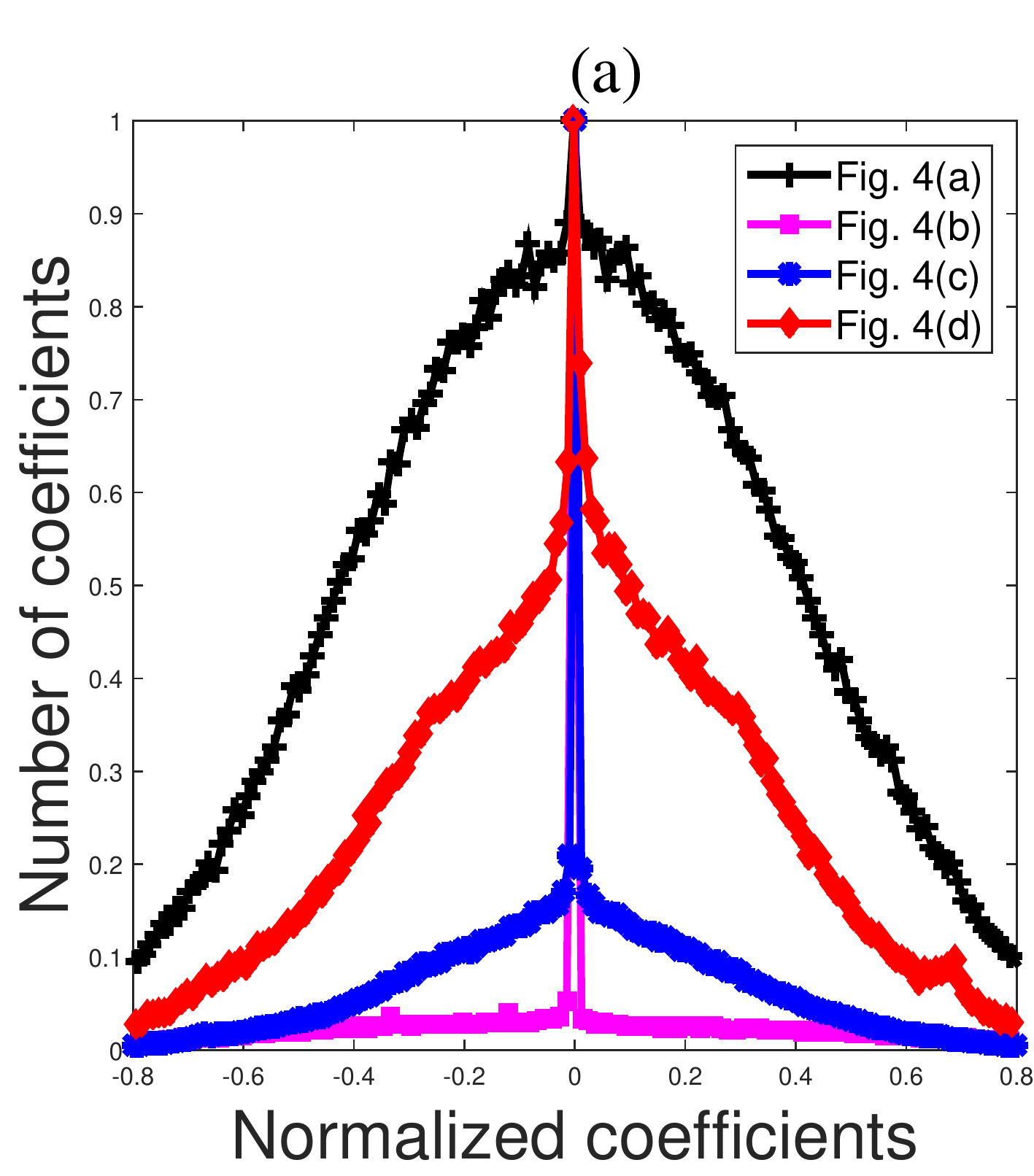} &
\includegraphics[width=1.9in]{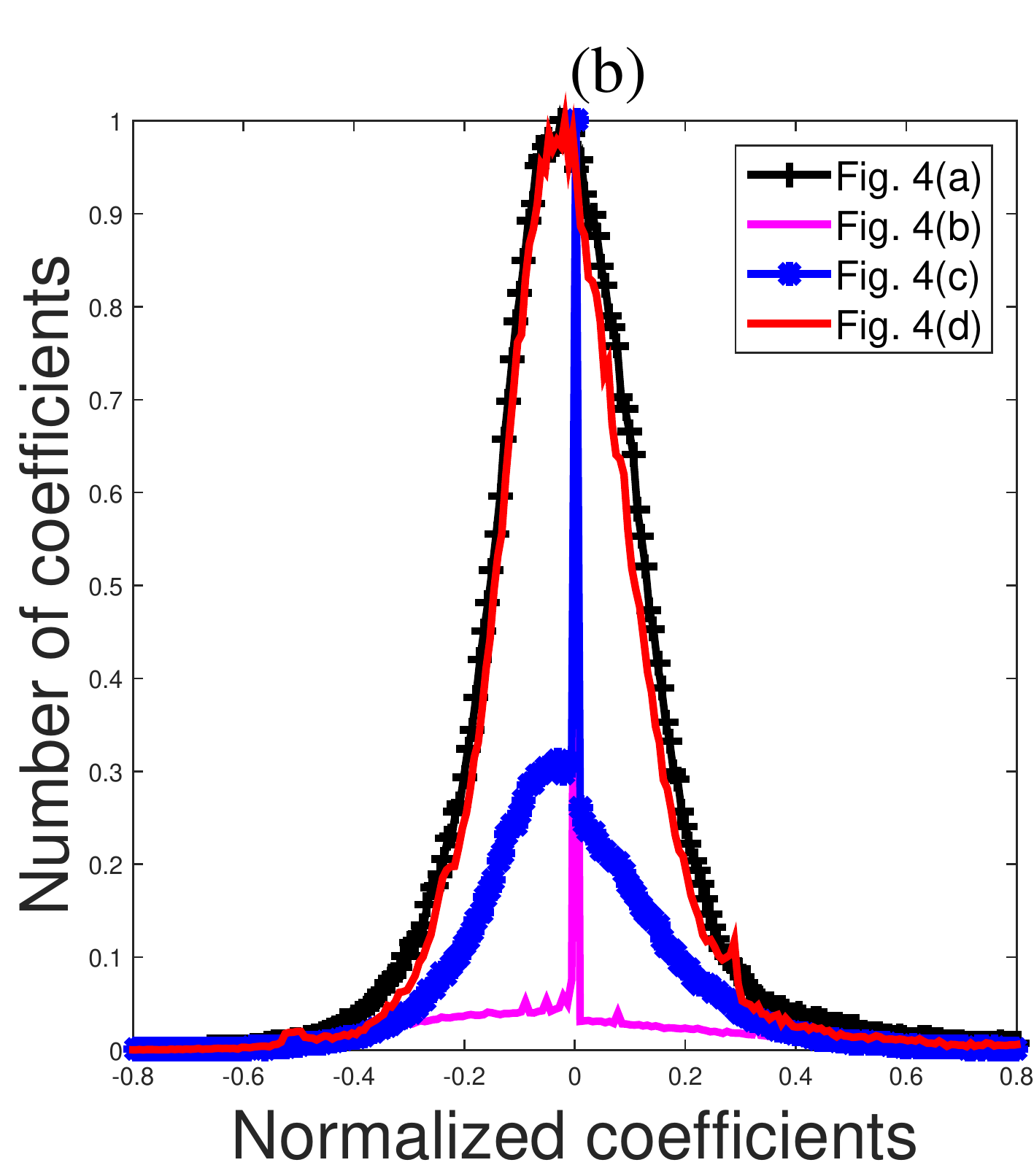} \\
\end{array}$
\caption{{Histogram of the normalized coefficients of (a) $Y$ and (b) $Y_{sigma}$ of Fig. \ref{fig:imgsForMaps} (a) - (d).}}
\label{fig:yellowFit}
\end{center}
\end{figure}
Divisive normalization of $\mathbf{Y}$ computed on a pristine image yields coefficients which, as illustrated in Fig. \ref{fig:yellowFit}(a), exhibit Gaussian-like behavior on good quality images. Furthermore, the normalized coefficients of the sigma map of $\mathbf{Y}$ (denoted henceforth as $\mathbf{Y_{sigma}}$) also display Gaussian behavior on pristine images (Fig. \ref{fig:yellowFit}(b)). This behavior is often not observed on distorted images. Thus, the goodness of generalized Gaussian fit of both the normalized coefficients of $\mathbf{Y}$ and $\mathbf{Y_{sigma}}$ at the original scale of the image are also extracted and added as features used in our model. As discussed in the next section, features drawn from the yellow color channel map was able to efficiently capture a few distortions that were not captured by the luminance component alone (Sec: Advantages of the proposed feature maps).

\subsection{Advantages of the proposed Feature Maps}
As an example to illustrate the advantages of the proposed feature maps, consider the four images presented in Fig. \ref{sampleImgs}. To reiterate, Fig. \ref{sampleImgs}(a) is a pristine image from the legacy LIVE Image Quality Database \cite{live-r2} while Fig. \ref{sampleImgs} (b) and (c) are JPEG2000 compression and additive white noise distortions (respectively) artificially applied to Fig. \ref{sampleImgs} (a). On the other hand, Fig. \ref{sampleImgs} (d) is a blurry image distorted by low-light noise and presumably compression, drawn from the LIVE In the Wild Image Quality Challenge Database \cite{crowdsource}.

\begin{figure}[t]
\begin{center}$
\begin{array}{ccc}
\includegraphics[width=1.9in]{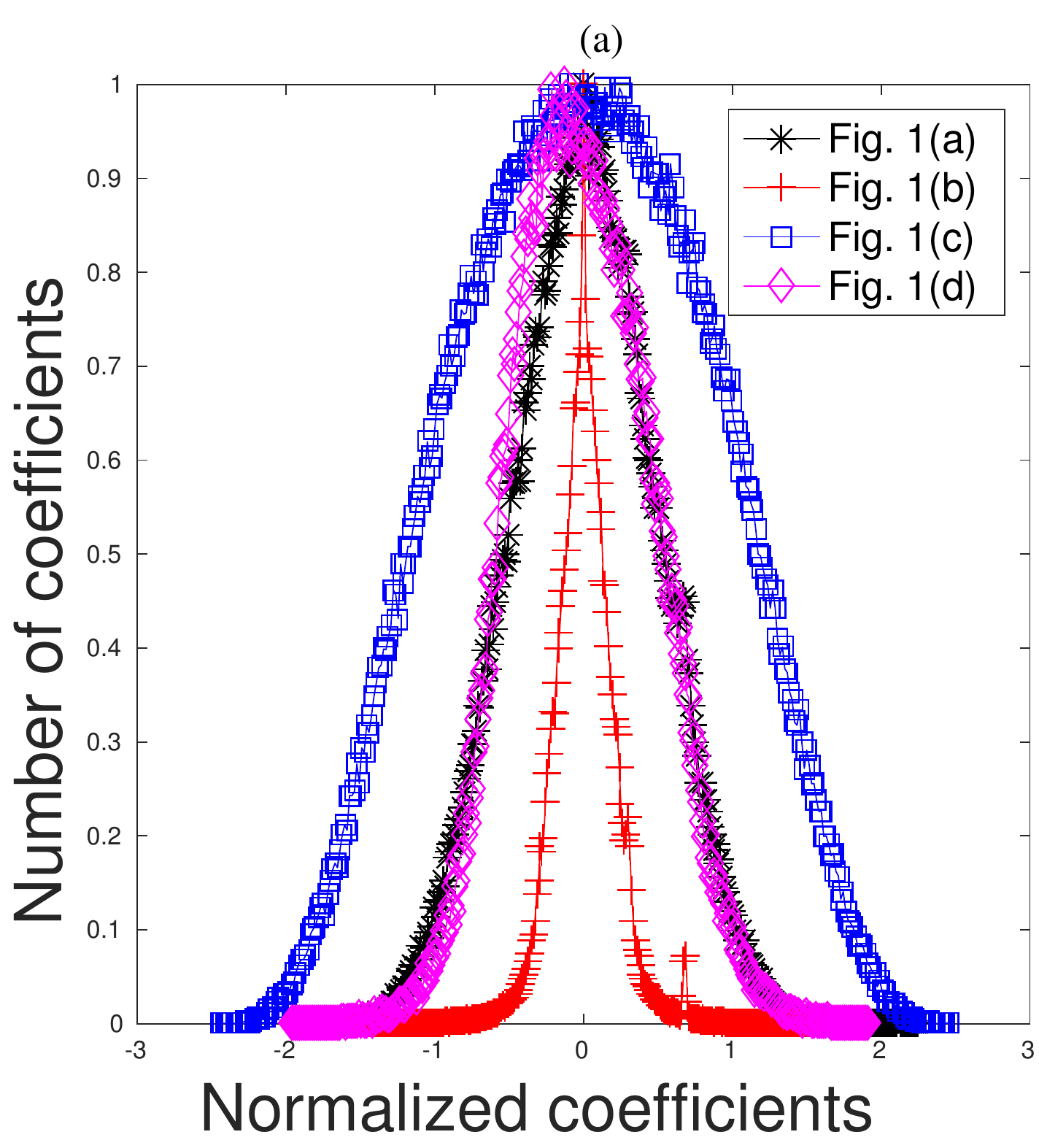} &
\includegraphics[width=1.9in]{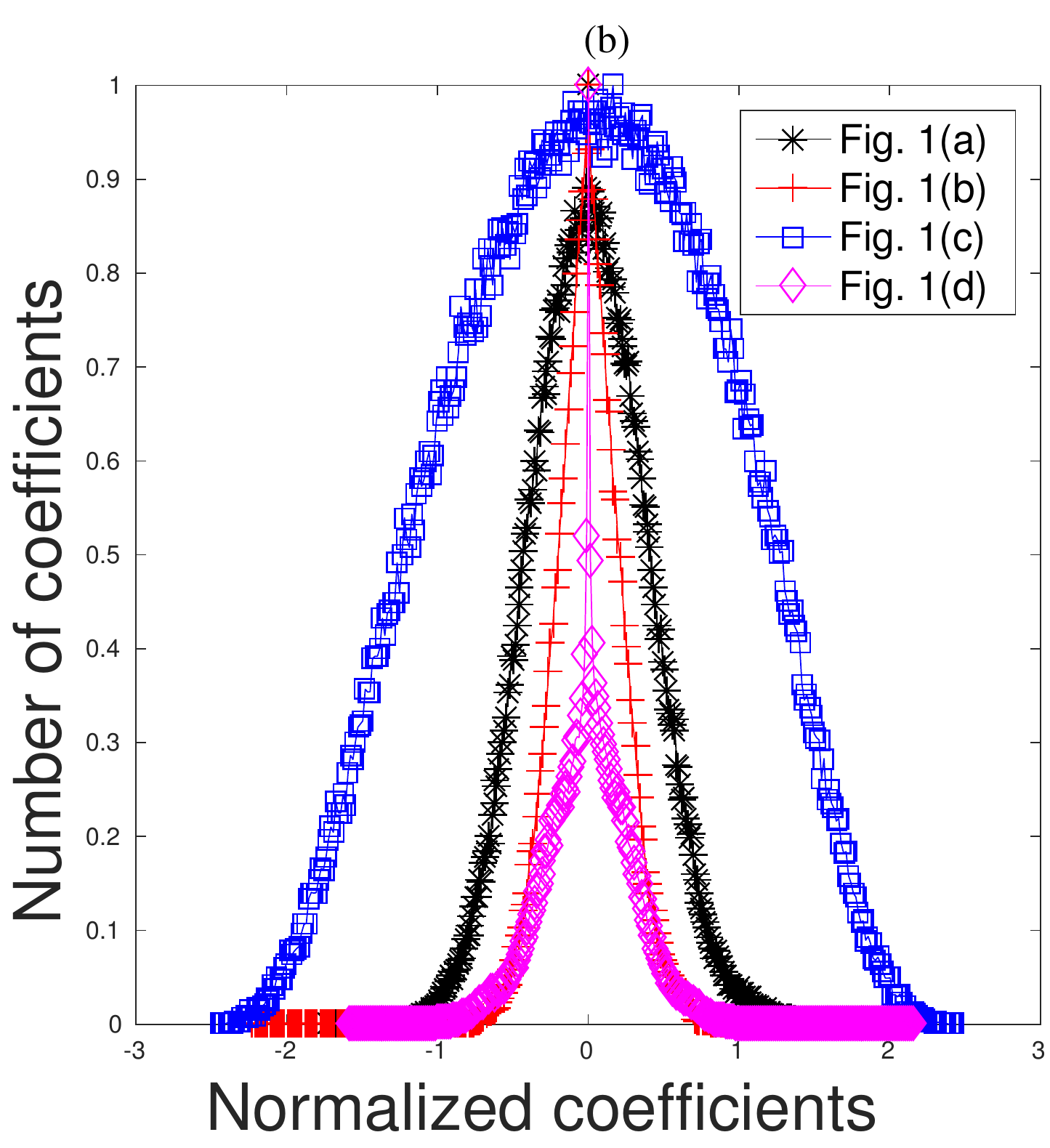} &
\includegraphics[width=1.9in]{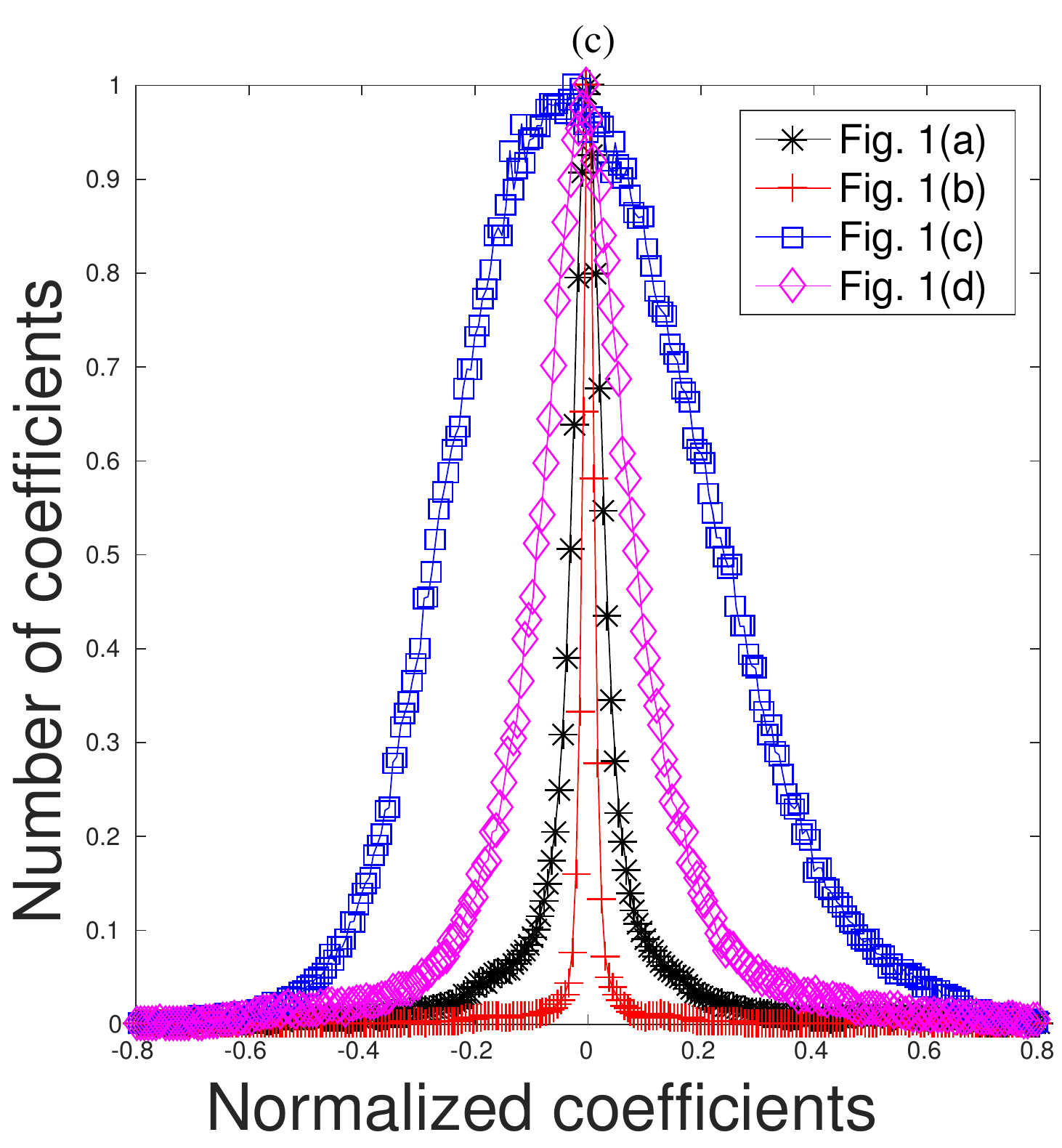} 
\\
\end{array}$
\caption{\small{(a) Histogram of the normalized coefficients of the images in Figures \ref{sampleImgs}(a) - (d) when processed using (a) BRISQUE-like normalization defined in (\ref{nlc_eqn}), (b) yellow color channel maps (\ref{yellowLedBetter}), and (c) $DoG\_sigma$. Notice how for the authentically distorted image Fig. \ref{sampleImgs} (d), the corresponding histogram in (a) resembles that of a pristine image. But in the case of the two feature maps - yellow color map and $DoG_{sigma}$, the histograms of pristine vs. authentically distorted images vary. (Best viewed in color).}}
\label{fig:compareNLC}
\end{center}
\end{figure} 

We processed these four images using three different operations - (a) the mean subtraction, divisive normalization operation used in \cite{brisque} on singly distorted images, (b) the yellow color channel map (\ref{yellowLedBetter}), and (c) the $DOG_{sigma}$ map defined in earlier sections. It may be observed that, though the histograms of the singly distorted images differ greatly from those of the pristine image in Fig. \ref{fig:compareNLC}(a), the distribution of an authentically distorted image containing noise, blur, and compression artifacts closely resembles the distribution of the pristine image. However, when the normalized coefficients of the proposed yellow color channel and the $DOG_{sigma}$ feature maps are observed in Fig. \ref{fig:compareNLC}(b)-(c), it is clear that these distributions are useful for distinguishing between the pristine image and both singly and authentically distorted images. We have observed the usefulness of all of the proposed feature maps on a large and comprehensive collection of images contained in the LIVE Challenge database.

We have thus far described a series of statistical features that we extract from a set of feature maps and also how each of these statistics are affected by the presence of image distortions (summarized in Table \ref{table:feature-all}\footnote{We did not show the 4 features extracted in the HSI space due to space constraints. The number of features shown in the LMS column refer to the sum of the number of features extracted on the $M$ and $S$ channel maps.}). Predicting the perceptual severity of authentic distortions is recondite, and a `bag of features' approach is a powerful way to approach the problem.

\subsection{Regression}
These perceptually relevant image features, along with the corresponding real-valued MOS of the training set, are used to train a support vector regressor (SVR). SVR is the most common tool for learning a non-linear mapping between image features and a single label (quality score) among IQA and VQA algorithms \cite{brisque, cdiivine, diivine, bliinds2, desique, todd}. Given an input image (represented by a feature vector), SVM maps the high-dimensional feature vector into a visual quality score \cite{dataMining, cckao}. While the database is large, it is not large enough to motivate deep learning methods. The SVM classifier and regressor is widely used in many disciplines due to its high accuracy, ability to deal with high-dimensional data, and flexibility in modeling diverse sources of data \cite{dataMining}.

In our algorithm, we used an SVR with a radial basis kernel function. Following this, given any test image's image features as input to the trained SVR, a final quality score may be predicted. The optimal model parameters of the learner were found via cross-validation. Our choice of the model parameters was driven by the obvious aim of minimizing the learner's
fitting error to the validation data. 

\section{Experiments}
In the following, we refer to our blind image quality assessment model as the \textbf{F}eature maps based \textbf{R}eferenceless \textbf{I}mage \textbf{QU}ality \textbf{E}valuation \textbf{E}ngine (FRIQUEE), following IQA naming conventions and to achieve brevity. FRIQUEE combines a large, diverse collection of perceptually relevant statistical features across multiple domains, which are used to train a regressor that is able to conduct blind image quality prediction. Variations called FRIQUEE-Luma, FRIQUEE-Chroma, FRIQUEE-LMS, and FRIQUEE-ALL are developed according to the subset of overall features considered. Thus FRIQUEE-Luma uses feature maps a. - f., FRIQUEE-Chroma uses feature maps g. - h., FRIQUEE-LMS uses feature maps i.-j., while FRIQUEE-ALL uses all feature maps as well as the two HSI color space feature maps and the yellow color channel map.

In all of the experiments we describe below, the model (initialized with the optimal parameters) was trained from scratch on a random sample of 80\% training images and tested on the remaining non-overlapping 20\% test data. To mitigate any bias due to the division of data, the process of randomly splitting the dataset was repeated 50 times. Spearman's rank ordered correlation coefficient (SROCC) and Pearson's correlation coefficient (PLCC) between the predicted and the ground truth quality scores were computed at the end of each iteration. The median correlation over these 50 iterations is reported. A higher value of each of these metrics indicates better performance both in terms of correlation with human opinion as well as the performance of the learner. We also report the outlier ratio (OR) \cite{outlier} which is the fraction of the number of predictions lying outside the range of $\pm$ 2 times the standard deviations of the ground truth MOS. A lower value of the outlier ratio indicates better performance of a given model. 

\subsection{Comparing Different IQA Techniques} \label{diff-iqa} 
We trained several other well-known NR IQA models\footnote{In the case of DIIVINE \cite{diivine} and C-DIIVINE \cite{cdiivine} which are two-step models, we skipped the first step of identifying the probability of an image belonging to one of the five distortion classes present in the legacy LIVE IQA Database as it doesn't apply to the newly proposed database. Instead, after extracting the features as proposed in their work, we learned a regressor on the training data.} (whose code was publicly available) on the LIVE In the Wild Image Quality Challenge Database, using identical train/test settings and the same cross-validation procedure over multiple random trials. An SVR with a RBF kernel was trained using FRIQUEE features and we denote this model as FRIQUEE-ALL. The median and the standard deviations of the correlations and the mean of the outlier ratios\footnote{The NIQE \cite{niqe} score is a measure of how far a given image is from `naturalness,' which is different from the subjective MOS values. Since it is not trained on MOS values, we do not compute an outlier ratio on the predicted NIQE scores.} across the 50 train-test iterations is reported in Table \ref{tbl:svr} from which we may conclude that the performance of the proposed model on unseen test data is significantly better than that of current top-performing state-of-the-art NR IQA models on the LIVE Challenge Database \cite{crowdsource, deepti-crowdsource}. 

To justify our design choice of an SVR with an RBF kernel, we also trained a linear SVR (FRIQUEE-LSVR) on FRIQUEE features extracted from the images in the LIVE Challenge Database. The training was performed under the same setting (on 50 random train/test splits). The median correlations across 50 iterations are reported in Table \ref{tbl:svr}. From this table we may conclude that the performance of FRIQUEE-ALL is better than the other learners. Also, comparing the median correlation scores of FRIQUEE-ALL with those of top-performing IQA models such as C-DIIVINE, BRISQUE, and DIIVINE, all of which also use an SVR as a learning engine, reveals that the perceptually-driven FRIQUEE NSS features perform better than the features designed in the other top-performing IQA models.

The high internal statistical consistency and reliability of the subjective scores gathered in the crowdsource study make it possible to consider the MOS values obtained from the online study as ground truth quality scores of the images \cite{deepti-crowdsource}. Moreover, the poor correlation scores reported by most algorithms suggests that the LIVE Challenge Database is a difficult test of the generalizability of those models.

\begin{table}[t] 
\centering
\caption{Median PLCC and SROCC, and mean OR of several no-reference IQA metrics across 50 train-test combinations on the LIVE Challenge database \protect\cite{crowdsource, deepti-crowdsource}. FRIQUEE-ALL refers to the scenario where the proposed learning engine, i.e.,SVR with an RBF was used.}
\vspace{0.5cm}
 \begin{tabular}{| >{\centering\arraybackslash}m{2.7in} | >{\centering\arraybackslash}m{1.2in} | >{\centering\arraybackslash}m{1.2in} | >{\centering\arraybackslash}m{1.2in} | }
    \hline
&  {PLCC} & {SROCC} & {OR}\tabularnewline
    \hline
FRIQUEE-ALL & $0.70 \pm  0.04$ & $0.66 \pm 0.04$ & $0.04$ \tabularnewline
\hline
FRIQUEE-LSVR & $0.65 \pm 0.04$ & $0.62 \pm 0.04$ & $0.04$ \tabularnewline
\hline
BRISQUE \cite{brisque}&  $0.61 \pm 0.06$ & $0.58 \pm 0.05 $ & $0.07$\tabularnewline
\hline
DIIVINE \cite{diivine}& $0.59 \pm 0.05$ & $0.56 \pm 0.05$ &    $0.06$ \tabularnewline
\hline
BLIINDS-II \cite{bliinds2} & $0.45 \pm 0.05$ & $0.40 \pm 0.05$ & $0.09$ \tabularnewline
\hline
NIQE \cite{niqe} & $0.48 \pm 0.05$ & $0.42 \pm 0.05$ & $-$ \tabularnewline 
\hline
C-DIIVINE \cite{cdiivine} & $0.66 \pm 0.04$ & $0.63 \pm  0.04$ & $0.05$ \tabularnewline
\hline
\end{tabular} \label{tbl:svr}
\vspace{0.5cm}
\end{table} 

\begin{table}[t] 
\centering
\caption{{Results of the paired one-sided t-test performed between SROCC values generated by different measures. `1,' `0,' `-1' indicate that the NR IQA algorithm in the row is statistically superior, equivalent, or inferior to the algorithm in the column.}}
\vspace{0.4cm}
 \begin{tabular}{|>{\centering\arraybackslash}m{0.7in} | >{\centering\arraybackslash}m{0.7in}|>{\centering\arraybackslash}m{0.7in}| >{\centering\arraybackslash}m{0.7in} | >{\centering\arraybackslash}m{0.7in} | >{\centering\arraybackslash}m{0.7in} | >{\centering\arraybackslash}m{0.7in} |>{\centering\arraybackslash}m{0.7in} |>{\centering\arraybackslash}m{0.7in} | }
  \hline
&DIIVINE & BRISQUE & {NIQE} & {C-DIIVINE} & {BLIINDS-II} & {FRIQUEE-LSVR} & {FRIQUEE-ALL} \tabularnewline
    \hline
{DIIVINE} & 0 & -1 & 1 & -1 & 1& -1 & -1 \tabularnewline
\hline
{BRISQUE}& 1 & 0 & 1 & -1  & 1 & -1 & -1\tabularnewline
\hline
{NIQE} & -1 & -1 & 0 & -1 & 0 & -1 & -1\tabularnewline 
\hline
C-DIIVINE & 1 & 1 & 1 & 0 & 1 & 0 & -1 \tabularnewline
\hline
BLIINDS-II & -1 & -1 & 0 & -1 & 0 & -1 &-1 \tabularnewline
\hline
FRIQUEE-LSVR & 1 & 1 & 1 & 0 & 1 & 0 & -1 \tabularnewline
\hline
FRIQUEE-ALL & 1 & 1 & 1 & 1 & 1 & 1 & 0\tabularnewline
\hline
\end{tabular} \label{tbl:statSign}
\vspace{0.4cm}
\end{table} 

\subsection{Statistical Significance and Hypothesis Testing}
Although there exist apparent differences in the median correlations between the different algorithms (Table \ref{tbl:svr}), we evaluated the statistical significance of the performance of each of the algorithms considered. Thus, we performed hypothesis testing based on the paired t-test \cite{tTest} on the 50 SROCC values obtained from the 50 train-test trials. The results are tabulated in Table \ref{tbl:statSign}. The null hypothesis is that the mean of the two paired samples is equal, i.e., \emph{the mean correlation for the (row) algorithm is equal to the mean correlation for the (column) algorithm with a confidence of 95\%}. The alternative hypothesis is that \emph{the mean correlation of the row algorithm is greater than or lesser than the mean correlation of the column algorithm}. A value of `1' in the table indicates that the row algorithm is statically superior to the column algorithm, whereas a `-1' indicates that the row is statistically worse than the column. A value of `0' indicates that the row and column are statistically indistinguishable (or equivalent), i.e., we could not reject the null hypothesis at the 95\% confidence level.

From Table \ref{tbl:statSign} we conclude that FRIQUEE-ALL is statistically superior to all of the no-reference algorithms that we evaluated, when trained and tested on the LIVE Challenge Database.

\begin{table}[t]
\centering
\caption{{Median PLCC and Median SROCC across 50 train-test combinations on \protect\cite{crowdsource, deepti-crowdsource} when FRIQUEE features from each color space were independently used to train an SVR.}}
\vspace{0.4cm}
 \begin{tabular}{| >{\centering\arraybackslash}m{1.5in} | >{\centering\arraybackslash}m{1.5in} | >{\centering\arraybackslash}m{1.5in} |  }\hline
&  {PLCC} & {SROCC}  \tabularnewline
    \hline
FRIQUEE-Luma & $0.64 \pm 0.04$ & $0.61 \pm 0.04$ \tabularnewline
\hline
FRIQUEE-LMS &  $0.63  \pm 0.04$ & $0.60 \pm 0.04 $ \tabularnewline
\hline
FRIQUEE-Chroma &  $0.36  \pm 0.05$ & $0.34 \pm 0.05$ \tabularnewline
\hline
\end{tabular}
\vspace{0.4cm}
 \label{colorspace} 
\end{table}

\subsection{Contribution of Features from Each Color Space} \label{sec:colorspace}
We next evaluated the performance of FRIQUEE-Luma, FRIQUEE-Chroma, and FRIQUEE-LMS. We trained three separate SVRs with features extracted from each color space serving as an input to each SVR and report the median correlation values across 50 random train/test splits in Table \ref{colorspace}. These values justify our choice of different color spaces, all of which play a significant role in enhancing image quality prediction. 

\subsection{Contribution of Different Feature Maps}
\begin{figure}[t] 
\begin{center}
\includegraphics[width=6in]{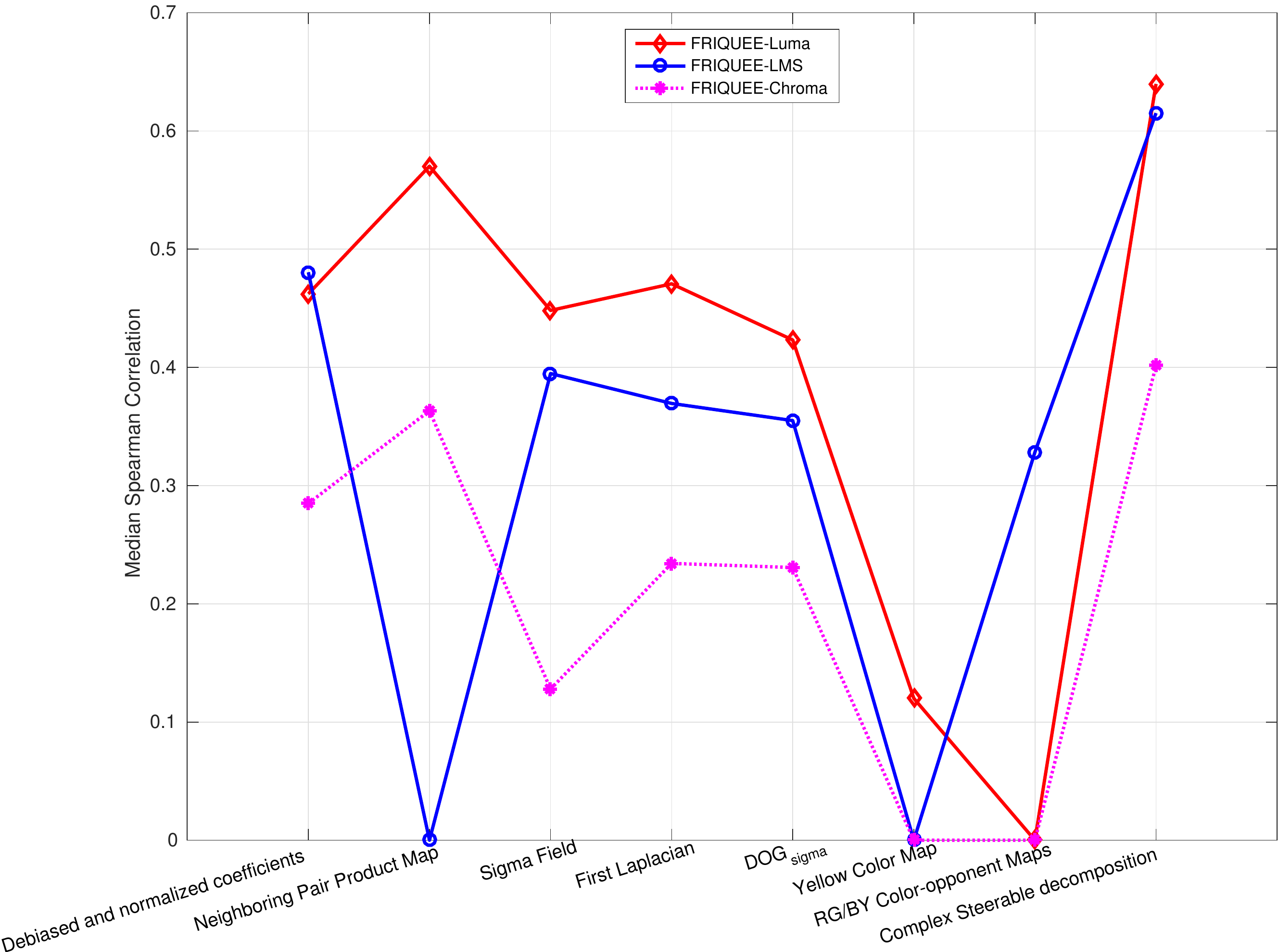}
\caption{{Contribution of different \emph{types} of features that are extracted in different color spaces. A correlation of 0 in a color space indicates that that specific feature map was not extracted in that color space.}}
\label{fig:corrPlot}
\end{center}
\end{figure}

To better understand the relationship between our feature set and perceptual quality, we trained separate learners (SVR with radial basis kernel functions) on the statistical features extracted from each feature map on 50 random, non-overlapping train and test splits. We report the median Spearman rank ordered correlation scores over these 50 iterations in Fig. \ref{fig:corrPlot}. This plot illustrates the degree to which each of these features accurately predict perceived quality, while also justifying the choice of the feature set \footnote{We included the Yellow Map under FRIQUEE-Luma in Fig. \ref{fig:corrPlot} purely for the purpose of illustration. It is not extracted from the luminance component of an image but is a color feature as described earlier.}.
\subsection{Evaluating the Robustness of Different IQA Techniques}\label{sec:robust} The goal of this experiment was to study the efficacy of training IQA models on the synthetically distorted images contained in current benchmark databases relative to training on authentically distorted images. Some of the current top-performing IQA learning models have been made publicly available (i.e., the model parameter values used by their SVRs) after being trained on the images on the legacy LIVE IQA Database. We sought to understand the performance of these publicly available models when they are used in real-world scenarios, to predict the quality of real-world images captured using mobile devices. We used the publicly available model BRISQUE \cite{brisque} trained on the legacy LIVE Database. With regards to the other blind algorithms, we extracted image features from each image in the LIVE IQA Database following the same procedure as was originally presented in their work and separately trained SVRs for each model on these image features. 

Next, we used the 50 randomly generated test splits and evaluated each learned engine (trained on the LIVE IQA Database) on the 50 test splits. We report the median of the correlations of the predicted scores with human judgments of visual quality across the 50 test splits in Table \ref{tbl:trainOnLegacy}. This analysis provides an idea of how well state-of-the-art quality predictors generalize with respect to image content and real-world distortions. As can be seen from the results reported in Table \ref{tbl:trainOnLegacy}, although FRIQUEE performed better than all of the algorithms, the performance of all the models suffer when they are trained only on images containing synthetic, inauthentic distortions.

\begin{table}[t] 
\centering
\caption{{Median PLCC and Median SROCC across 50 train-test combinations of a few NR-IQA models on \protect\cite{crowdsource, deepti-crowdsource} when models trained on the LIVE IQA database are used to predict the quality of the images from the LIVE Challenge database.}}
\vspace{0.4cm}
 \begin{tabular}{| >{\centering\arraybackslash}m{2.7in} | >{\centering\arraybackslash}m{1.1in} | >{\centering\arraybackslash}m{1.1in} |  }
    \hline
&  {PLCC} & {SROCC} \tabularnewline
    \hline
FRIQUEE-ALL &  $0.6289 \pm 0.0425$ & $0.6303 \pm 0.0405$ \tabularnewline
\hline
BRISQUE \cite{brisque}& $0.3296 \pm 0.0505 $ & $0.2650 \pm 0.0505$ \tabularnewline
\hline
DIIVINE \cite{diivine}&  $0.3667 \pm  0.0504$ & $0.3328 \pm 0.0536$ \tabularnewline
\hline
BLIINDS-II \cite{bliinds2} & $0.1791 \pm 0.0713 $ & $0.1259  \pm 0.0704$\tabularnewline
\hline
C-DIIVINE \cite{cdiivine} &  $0.4705 \pm 0.0549$ & $0.4589 \pm 0.0515$ \tabularnewline
\hline
\end{tabular} \label{tbl:trainOnLegacy}
\vspace{0.4cm}
\end{table}

\subsection{Evaluating IQA models on Legacy LIVE Database}\label{sec:live-r2}
We next compared the performance of our model against several other top-performing blind IQA models on the older standard benchmark LIVE IQA Database \cite{live-r2}. Regarding FRIQUEE-ALL, 564 features were extracted on all the images of the LIVE IQA Database and the image data was divided into training and test subsets, with no overlap in content. This process was repeated 1000 times and we report the median correlation values in Table \ref{legacy-results}. With regards to the other models, we report the median correlation scores as reported in their papers. We note that \cite{tang-cvpr} report an SROCC value of 0.9650 on the legacy LIVE database, but this result is not verifiable since the authors do not make the code publicly available. Since we cannot validate their claim, we do not include it in Table \ref{legacy-results}.
\begin{table}[t]
\centering
\caption{{Performance on legacy LIVE IQA database \protect\cite{live-r2}. Italics indicate NR-IQA models. -NA- indicates data not reported in the corresponding paper.}}
 \begin{tabular}{| >{\centering\arraybackslash}m{2.8in} | >{\centering\arraybackslash}m{1.3in} | >{\centering\arraybackslash}m{1.3in} |}
\hline
& {SROCC} & {PLCC} \tabularnewline
\hline
PSNR & 0.8636 & 0.8592 \tabularnewline
\hline
SSIM \cite{ssim} & 0.9129 & 0.9066 \tabularnewline
\hline
MS-SSIM \cite{ms-ssim} & 0.9535 & 0.9511 \tabularnewline
\hline
\textit{CBIQ} \cite{cbiq} & \textit{0.8954} & \textit{0.8955} \tabularnewline
\hline
\textit{LBIQ} \cite{lbiq} & \textit{0.9063} & \textit{0.9087}\tabularnewline
\hline
\textit{DIIVINE} \cite{diivine} & \textit{0.9250} & \textit{0.9270} \tabularnewline
\hline
\textit{BLIINDS-II} \cite{bliinds2} & \textit{0.9124} & \textit{0.9164} \tabularnewline
\hline
\textit{BRISQUE} \cite{brisque} & 0.9395 & \textit{0.9424} \tabularnewline
\hline
\textit{NIQE} \cite{niqe} & 0.9135 & 0.9147 \tabularnewline
\hline
\textit{C-DIIVINE} \cite{cdiivine} & 0.9444 & 0.9474 \tabularnewline
\hline
\textit{FRIQUEE-ALL} & \textit{0.9477} $\pm 0.0250$ & \textit{\textbf{0.9620} $\pm 0.0223$} \tabularnewline
\hline
 \end{tabular}
\vspace{0.1cm}
\label{legacy-results}
\end{table}
\begin{table}[t]
\centering
\caption{{Median PLCC and Median SROCC across 100 train-test combinations of a few NR-IQA models on LIVE-Multiply Database - Part I \protect\cite{liveMultiply}.}}
\vspace{0.4cm}
 \begin{tabular}{| >{\centering\arraybackslash}m{2.7in} | >{\centering\arraybackslash}m{1.1in} | >{\centering\arraybackslash}m{1.1in} |  }
    \hline
&  {PLCC} & {SROCC} \tabularnewline
    \hline
FRIQUEE-ALL &  $\textbf{0.9667}$ & $\textbf{0.9591}$ \tabularnewline
\hline
BRISQUE \cite{brisque}& $0.9391$ & $0.9238$ \tabularnewline
\hline
DIIVINE \cite{diivine}&  $ 0.9424$ & $ 0.9327$ \tabularnewline
\hline
NIQE \cite{niqe} & $0.9075$ & $0.8614$\tabularnewline
\hline
C-DIIVINE \cite{cdiivine} &  $0.9336$ & $0.9179$ \tabularnewline
\hline
\end{tabular} \label{tbl:liveP1}
\vspace{0.4cm}
\end{table}
\begin{table}[t]
\centering
\caption{{Median PLCC and Median SROCC across 100 train-test combinations of a few NR-IQA models on LIVE-Multiply Database - Part II \protect\cite{liveMultiply}.}}
\vspace{0.4cm}
 \begin{tabular}{| >{\centering\arraybackslash}m{2.7in} | >{\centering\arraybackslash}m{1.1in} | >{\centering\arraybackslash}m{1.1in} |  }
    \hline
&  {PLCC} & {SROCC} \tabularnewline
    \hline
FRIQUEE-ALL &  $\textbf{0.9664}$ & $\textbf{0.9632}$ \tabularnewline
\hline
BRISQUE \cite{brisque}& $0.9070$ & $0.8748$ \tabularnewline
\hline
DIIVINE \cite{diivine}&  $ 0.8956$ & $ 0.8677$ \tabularnewline
\hline
NIQE \cite{niqe} & $0.8316$ & $0.7762$\tabularnewline
\hline
C-DIIVINE \cite{cdiivine} &  $0.8837$ & $0.8772$ \tabularnewline
\hline
\end{tabular} \label{tbl:liveP2}
\vspace{0.4cm}
\end{table}

\begin{table}[t]
\centering
\caption{{Median PLCC and Median SROCC across 100 train-test combinations of a few NR-IQA models on TID2013 Database \protect\cite{tid13}.}}
\vspace{0.4cm}
 \begin{tabular}{| >{\centering\arraybackslash}m{2.7in} | >{\centering\arraybackslash}m{1.1in} | >{\centering\arraybackslash}m{1.1in} |  }
    \hline
&  {PLCC} & {SROCC} \tabularnewline
\hline
FRIQUEE-ALL &  $\textbf{0.9287}$ & $\textbf{0.9138}$ \tabularnewline
\hline
BRISQUE \cite{brisque}& $0.7781$ & $0.7515$ \tabularnewline
\hline
DIIVINE \cite{diivine}&  $ 0.8066$ & $ 0.7644$ \tabularnewline
\hline
NIQE \cite{niqe} & $0.3592$ & $0.3137$\tabularnewline
\hline
C-DIIVINE \cite{cdiivine} &  $0.7319$ & $0.6602$ \tabularnewline
\hline
\end{tabular} \label{tbl:tid}
\vspace{0.4cm}
\end{table}

\begin{table}[t]
\centering
\caption{{Median PLCC and Median SROCC across 100 train-test combinations of a few NR-IQA models on CSIQ Database \protect\cite{csiq}.}}
\vspace{0.4cm}
 \begin{tabular}{| >{\centering\arraybackslash}m{2.7in} | >{\centering\arraybackslash}m{1.1in} | >{\centering\arraybackslash}m{1.1in} |  }
    \hline
&  {PLCC} & {SROCC} \tabularnewline
    \hline
FRIQUEE-ALL &  $\textbf{0.9622}$ & $\textbf{0.9627}$ \tabularnewline
\hline
BRISQUE \cite{brisque}& $0.8926$ & $0.8823$ \tabularnewline
\hline
DIIVINE \cite{diivine}&  $ 0.9171$ & $ 0.9282$ \tabularnewline
\hline
NIQE \cite{niqe} & $0.6943$ & $0.6142$\tabularnewline
\hline
C-DIIVINE \cite{cdiivine} &  $0.8660$ & $0.8611$ \tabularnewline
\hline
\end{tabular} \label{tbl:csiq}
\vspace{0.4cm}
\end{table}
Comparing the correlation scores reported in Table \ref{tbl:svr} with those in Table \ref{legacy-results}, we observe that several other blind IQA models are not robust to authentic distortions, since while they achieve superior performance on the legacy LIVE database, they fail to accurately predict the quality of authentically distorted images. On the other hand, it may be observed that FRIQUEE not only performs well on the LIVE Challenge database (Table \ref{tbl:svr}), but also competes very favorably with all the other blind IQA models as well as with full-reference IQA models on the legacy LIVE Database. It reaches and exceeds the `saturation level' of performance achieved on this long-standing synthetic distortion database by the tested prior models. This supports our contention that a combination of semantically rich, perceptually informative image features feeding a highly discriminative learning model is a powerful way to automatically predict the perceptual quality of images afflicted by both authentic and synthetic distortions.

\subsection{Evaluating IQA models on other legacy database}\label{sec:allDBs}
Although our primary focus was to evaluate the performance of our proposed algorithm on the LIVE In the Wild Challenge Database (since we wanted to benchmark the superior performance of FRIQUEE on authentically distorted images), we understand that some readers may find performance on the legacy databases to be relevant. Therefore, we evaluated FRIQUEE and a few other top-performing NR IQA algorithms on other legacy databases such as TID2013 \cite{tid13} and CSIQ \cite{csiq} both of which contain single, synthetic distortions and LIVE-Multiply Database \cite{liveMultiply}, which contains Gaussian blur followed by JPEG compression distortions (in Part I) and Gaussian blur followed by additive white noise distortions (in Part II). The images in all of these datasets were divided into non-overlapping training and test sets and this process was repeated 100 times. For each IQA algorithm on every database, optimal model parameters were chosen for an SVR with a radial basis kernel while training a model. In Tables \ref{tbl:liveP1} - \ref{tbl:csiq}, we report the median correlation values between ground truth and predicted quality scores across 100 iterations on all these databases.

Comparing the correlation scores, it may be observed that FRIQUEE features perform better than the features designed in all the other top-performing IQA models on synthetic distortions modeled in \cite{tid13, csiq, liveMultiply}.

\section{Conclusions and Future Work}
We have described a first effort towards the design of blind IQA models that are capable of predicting the perceptual quality of images corrupted by complex mixtures of authentic distortions. Its success encourages us to explore the feasibility of developing analogous powerful blind \emph{video} quality assessment models using space-time natural video statistics models \cite{live-v1, live-v2, live-v3}, and also to practically adapt our model for application to real-world problems, such as perceptual optimization of digital camera capture.

\bibliography{friquee_jov_arxv}
\bibliographystyle{jovcite}

\end{document}